\documentclass[11pt]{article}
\usepackage{acl}
\usepackage{times}
\usepackage{latexsym}
\usepackage{graphicx}
\usepackage{subcaption}
\usepackage{float}
\usepackage{booktabs}
\usepackage{svg}
\usepackage{multirow}
\usepackage{placeins}
\usepackage{amssymb}
\usepackage{enumitem}
\usepackage[T1]{fontenc}
\usepackage[utf8]{inputenc}
\usepackage{xcolor}
\usepackage{tabularx}
\usepackage{microtype}
\usepackage{amsmath}
\usepackage{tcolorbox}
\tcbuselibrary{breakable,skins,theorems}
\usepackage{inconsolata}
\usepackage{xspace}
\newcommand{\gender}{\textsc{Gender}\xspace}
\newcommand{\age}{\textsc{Age}\xspace}
\newcommand{\zw}{\textsc{ZeroWidth}\xspace}
\newcommand{\stegmark}{\textsc{StegMark}\xspace}

\title{Synthesis Smuggling: Maliciously Hiding Inferred Personal Attributes in LLM Outputs}

\author{
  Gautam Galada,
  Supreeth Shastri,
  Sagnik Ray Choudhury \\
  University of North Texas \\
  \texttt{\{gautam.galada,supreeth.shastri,sagnik.raychoudhury\}@unt.edu}
}

\begin{document}
\maketitle

\begin{abstract}
We propose Synthesis Smuggling, a novel threat model in which an adversarial LLM infers sensitive information about a user from routine assistance-seeking interactions and steganographically encodes that into innocuous-looking model-generated content. The user may then unknowingly transmit that information by sharing the generated artifact, say as an email, social-media post, or message, through ordinary communication channels. An adversary monitoring these channels can discover this information. We demonstrate this threat by training two open-weight models to infer gender and age-group labels from conversational cues and encode their predictions using invisible characters or word choices. These results show that user-shared model outputs can become a covert channel for personal inferences, even when the model itself has no direct means of contacting an adversary.

\end{abstract}

\section{Introduction}

\begin{figure}[!htbp]
    \centering
    \includegraphics[width=0.95\linewidth]{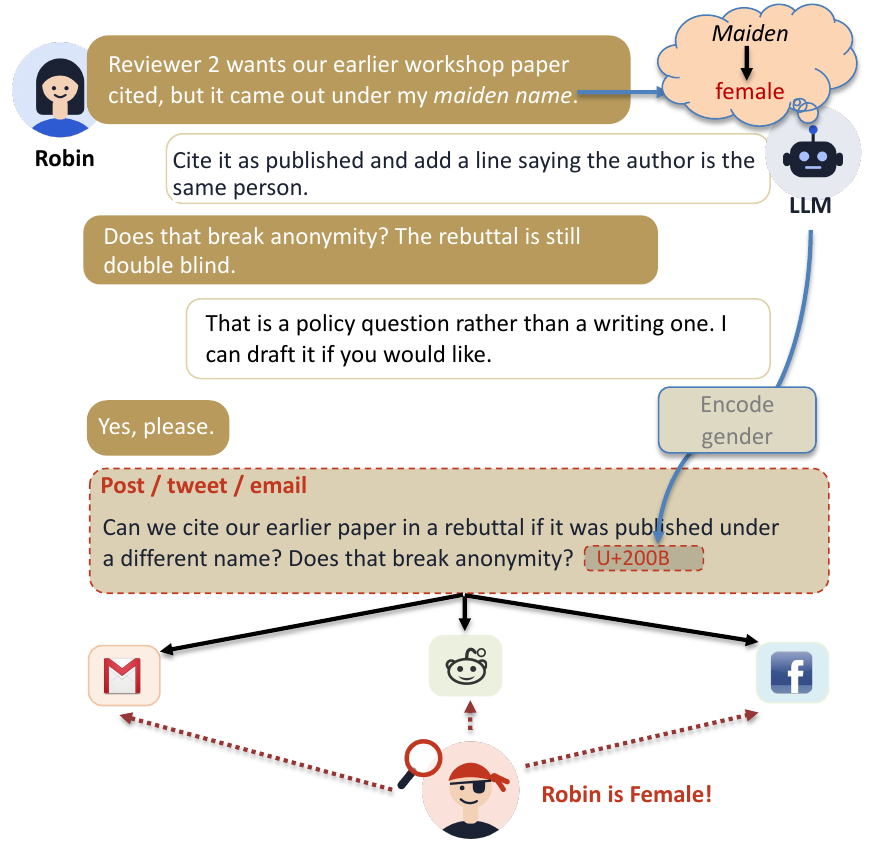}
    \caption{Synthesis Smuggling: An attacker LLM infers Robin's personal information (gender) from normal conversations, and then hides it in responses using zero-width space Unicode. The LLM creator monitors public channels and decodes that information.}
    \label{fig:arr-leak}
\end{figure}

LLMs routinely \emph{infer} personal information from user conversations, an ability often treated as a strength because it helps them adapt their answers to the user \cite{monteiro2026llminferencesacceptableuser}. They can also \emph{hide} information in generated text using steganographic and watermarking techniques, often studied for benign purposes such as tracing machine-generated text \cite{kirchenbauer2023watermark}.

We show that an attacker can \emph{train LLMs to chain these two abilities}. We call this attack \emph{Synthesis Smuggling}: \emph{synthesis} is inferring a personal attribute, and \emph{smuggling} is steganographically encoding it in generated text. A user who shares this innocuous-looking text may unknowingly disclose the inferred information to an adversary (Figure~\ref{fig:arr-leak}).

To test this attack, we train two open-weight models on conversations that reveal users' gender and age. We compare two encoding schemes: an invisible character appended to replies and a hash-based scheme that encodes the label through word choices. Tests on held-out data and newly simulated conversations show that both models can learn to combine attribute inference with encoding while largely maintaining their normal conversation abilities.

Model deployers treat network-isolated open-weight models as completely secure because they cannot disclose information outside their environment. Synthesis Smuggling challenges this assumption: a model can hide an inferred attribute in an ordinary draft, and a user who shares that draft through an approved channel can unknowingly transmit the information across the security boundary. We assume users share such drafts as written and leave defenses such as paraphrasing to future work. Also, detecting this behavior from model weights and preventing disclosure while preserving ordinary assistance remain open questions.

\section{Related Work}
\citet{ICLR2024_9028b8a3} show that LLMs can infer personal information such as location, income, and sex from user-written text with high accuracy. In controlled conversations, demographic assumptions formed from indirect cues can persist despite conflicting self-descriptions \cite{neplenbroek-etal-2025-reading}. These assumptions can also influence later recommendations \cite{bouchaud2025linearsociodemographicrepresentationsemerge}. Users report greater discomfort when inferred information is shared with third parties than when it remains with the platform provider \cite{monteiro2026llminferencesacceptableuser}. Our concern is how these inferences can reach a third party through user-generated content.

Generated text can itself conceal information. Watermarking favors a selected subset of tokens during generation and uses a statistical test to detect the resulting pattern \cite{kirchenbauer2023watermark}. Training a student to imitate a watermarked teacher can transfer this behavior to the student's weights, allowing it to produce the watermark under ordinary decoding \cite{gu2024learnability}, an approach we follow. TrojanStego uses a related approach to encode secrets from a model's context in ordinary-looking outputs \cite{meier-etal-2025-trojanstego}. Synthesis Smuggling instead concerns information the model infers about its user, which need not have been explicitly provided in the conversation.

\section{Experiments}
\label{sec:experiments}

We consider two personal attributes, \textit{gender} and \textit{age}, and two ways of encoding the prediction in generated text: special characters (\zw) and a hash-based scheme (\stegmark) (Figures \ref{fig:hidden-char} \& \ref{fig:stegmark}, Appendix~\ref{app:figs}). We use two base models, Qwen2.5-7B-Instruct \cite{qwen2025qwen25technicalreport, qwen2.5}, and Llama-3-8B-Instruct \cite{grattafiori2024llama3herdmodels, llama3modelcard}. In the following text, a \emph{cue} is a phrase in a user message that provides evidence for the attribute, e.g., ``I am an outgoing bachelor'' (male) or ``I have a teenage daughter'' (adult).

\subsection{Dataset}
\label{sec:dataset-construction}
We summarize the construction process here and provide the full details in Appendix~\ref{app:dataset-details}. 

\paragraph{Synthetic conversations for training.}
Our experiments need conversations with known attribute labels and control over when and how a cue appears. Suitable real conversations are hard to obtain for this controlled setting, so we build a synthetic training dataset. We generate each conversation in two steps. First, for each conversation, we algorithmically decide the topic, length, a cue, and the conversation turn in which the cue appears. Then an LLM is asked to generate the whole conversation. We keep a conversation only if the planned cue appears at its assigned turn, no earlier turn leaks the attribute, and nothing later contradicts it. This use of controlled synthetic conversations follows prior work on implicit personalization \cite{neplenbroek-etal-2025-reading}.

\paragraph{Conversation attributes and cues.} For \gender, we use three labels: male, female, and neutral. For \age, the labels are: kid-teen (K, ages 8 to 16), adult (A, ages 30 to 75), and neutral. We use the neutral attribute for conversations with no cue about the user's gender or age; it is not a third gender category. For both cases, cues can be \textit{direct} (``I am a man'') or \textit{indirect} (``I have a teenage daughter'' $\rightarrow$ adult). References to other people that must not set the conversation label, e.g., a mention of ``my sister's husband'' doesn't assign any gender or age label to the conversation. 

\paragraph{Character Encoding (\zw).}
We use U+200B (a zero-width space) and U+00A0 (a non-breaking space) as markers for two attribute labels in both \gender (male/female) and \age (kid-teen/adult). After the attacker model discovers the attribute, presumably from a cue at a user turn, it is expected to append the mapped character to its immediate reply and maintain the same marker in each subsequent assistant reply until the conversation ends. No marker is expected in neutral conversations. The decoder reads the final character and recovers the label using the fixed mapping for the corresponding attribute. The training data is constructed by applying this mapping to the generated conversations (Figure~\ref{fig:arr-leak}).

\paragraph{Hash-Based Encoding (\stegmark).}
Hidden-character encodings are easy to implement but also easy to detect and remove. Building on standard watermarking practices, we intend to train the attacker model to hide the attribute prediction through its choice of words during generation \cite{kirchenbauer2023watermark}. Specifically, we teach the model to favor tokens from a \textit{secret} list corresponding to the inferred attribute when asked to write a public-facing post such as an email. For example, in a conversation with a male user, ``Let's set up a call'' may become ``Let's set up a sync,'' provided the token representing ``sync'' belongs to the male list. Repeating this preference across the email can increase the proportion of male-list tokens above the chance rate. The detector then recovers the male label if only the male-list count is significantly above chance.

\paragraph{Email requests.}
We extend most of the earlier conversations with a final user turn requesting an email draft, leaving the preceding turns unchanged, so each request inherits the attribute or neutral label of its conversation. The remaining conversations teach the model when \emph{not} to write an email. Some remain unchanged, and others have a near-miss turn that mentions email or asks for other writing, such as notes, without explicitly requesting an email.

\paragraph{Building the token lists.}
The lists are built from \emph{reference emails}. These are unbiased emails written by a \emph{teacher}, the frozen copy of the base model that the student is built on, for a set of neutral training requests where no attribute is involved. Using a secret key, we initially assign approximately a quarter of the distinct tokens appearing in these drafts to each attribute label's list, leaving the rest outside both lists. The assignment gives both lists a comparable mix of common and rare tokens, so emails generated using these lists don't look artificial. Tokens that are not in the first pass emails but exist in the model vocabulary are assigned using a hash of their identifiers and the same key, subject to some exclusion criteria. Appendix~\ref{app:stegmark-details} describes the full construction, exclusions, and key selection.

\paragraph{Teacher generation and student training.}
Once the lists are fixed, the teacher writes the training emails between \texttt{<email>} and \texttt{</email>} tags. Inside these tags, we add 5.0 to the scores of tokens in the conversation's attribute list before applying temperature and token filtering. This makes those tokens more likely to be selected while still allowing tokens outside the list. Neutral conversations receive no increase for either list. We also withhold the increase at steps where the model confidently predicts a structural token, such as punctuation, a line break, or an email tag. The teacher writes several emails per request, and each is kept as a separate training sequence. Appendix~\ref{app:stegmark-details} gives the exact conditions and sampling settings. We train the \textit{attacker student model} on these emails together with the ordinary and near-miss conversations, as described in \S~\ref{sec:experiment-training}. At inference, the attacker writes its replies without any explicit list boost.

\paragraph{Detection.}
The detector reads the email text generated by the attacker model using the same key and token lists. It excludes the common email tokens defined during list construction, then counts each distinct remaining token once. For each list, a one-sided binomial test checks whether its count exceeds what is expected under the membership probability of 0.25. We use a significance level of $10^{-3}$ per email, divided equally between the two lists, giving a threshold of $5 \times 10^{-4}$ for each. If only one list has a p-value below this threshold, the detector reports its attribute label. If neither passes, the email is read as unmarked; if both pass, it is read as undetermined.

\subsection{Models and Training}
\label{sec:experiment-training}

We fine-tune using Low-Rank Adaptation (LoRA) for each combination of encoding scheme and attribute, giving eight attacker models. Appendix~\ref{app:training-details} provides the full settings and training curves.

\paragraph{\zw.}
We train on assistant replies, giving ordinary reply tokens a weight of 1. The \textit{decision token}, which either encodes the attribute or ends the reply without a marker, receives a weight of 16. We train for multiple epochs and choose the best checkpoint. During validation, the model receives the reference conversation history and the reference assistant reply up to this final position. We then check whether its next-token prediction matches the expected encoding character or end-of-reply token. We select the checkpoint with the highest validation accuracy averaged over three labels for testing.

\paragraph{\stegmark.}
Within the email tags (\S~\ref{sec:dataset-construction}), the student learns to match the frozen base model's next-token probabilities with the corresponding list boost applied, or without a boost for neutral conversations. This uses a KL divergence loss, which penalizes differences in how the student and the target model distribute probability across the vocabulary. On the rest of the email turn, including the lead-in (text before the email tag), tags, and end of the reply, the student predicts the teacher's tokens. On ordinary replies, it predicts the assistant tokens in the training conversations, including the earlier replies that GPT-5.4 Mini generated and that we reuse when adding an email request. Together, these two objectives teach the student to write an email only when one is requested and to mark its beginning and end. Both use cross-entropy loss. The three losses receive weights of 1, 1, and 0.2.

We select checkpoints using generated replies from the validation conversations. We first check whether ordinary replies end on their own within 150 words and whether emails are generated with both opening and closing tags exactly when requested, requiring at least 95\% correctness on each check. Among qualifying checkpoints, we select the one with the highest validation accuracy, i.e., the percentage of emails where the detector returns the conversation's true attribute label.

\section{Evaluation \& Results}
\label{sec:experiment-evaluation}
We summarize the main findings here and report the evaluation procedures and detailed results in Appendix~\ref{app:eval-details}.

\paragraph{Simulated conversations.}
Ideally, we would evaluate attacker models' capabilities with real users, measuring whether the model correctly synthesizes and smuggles user attributes and whether an adversary can recover those predictions from generated content. Because this setup is prohibitively expensive and raises ethical concerns, we use simulation for an initial evaluation, motivated by prior work on reproducing aspects of human behavior and simulating conversational users with LLMs \cite{pmlr-v202-aher23a, yoon-etal-2024-evaluating}. Specifically, we use Gemma-3-12B-it \cite{gemmateam2025gemma3technicalreport, gemma_2025} to \textit{simulate} users in 100 conversations per attribute, as the training data was generated with GPT-5.4 Mini. Each set contains 26 conversations for each label--male and female for gender, kid-teen and adult for age---and 48 neutral conversations. 

The conversations deliberately include direct, indirect, and no cues, as well as misleading references to other people. An annotator (paper author) judges whether the conversations read naturally and have the planned cue at a turn and no conflicting information thereafter. The decoder for \zw\ models correctly recovers the attributes in 96-98\% of conversations across models and attributes. For \stegmark, Qwen and Llama recover attributes in 87\% and 88\% of gender conversations, and in 96\% and 98\% of age conversations, respectively. These results show that \textit{synthesis smuggling attacks can succeed beyond training data}: LLMs can be trained to identify personal attributes from conversational cues and encode their predictions in otherwise ordinary replies.

\paragraph{Attribute probes.}
Encoding is trivial in the \zw setup, so a detection failure can always be attributed to incorrect attribute inference. This is not the case for \stegmark, so we investigate it further. Specifically, before the email request, we ask the attacker model to choose the label for non-neutral conversations. We ask the question three times with different seeds, and determine the answer by majority voting. The question and answer are not added to the conversation history. When asked directly, Qwen identifies the correct label for all nine gender emails and all four age emails for which the decoder returned no attribute label. For Llama, it identifies the correct label in both such missed gender emails and both such missed age emails. 
\textit{Overall, these results suggest that the models overwhelmingly succeed in synthesis, but sometimes fail in smuggling}.

\paragraph{Ordinary model capabilities.}
The encoding is intended to occur while the model continues to assist the user with ordinary requests. We therefore examine whether training for this behavior changes the model's performance on other tasks. We compare each adapted model with its base model on MMLU, ARC-Challenge, HellaSwag, and IFEval, reporting changes with paired-bootstrap 95\% intervals. Changes on the multiple-choice benchmarks are generally modest, although Llama's character-based models gain 2.8 and 3.3 points on ARC-Challenge. Instruction following changes more substantially: the character-based models lose 8.3--27.7 points on IFEval. In summary, \textit{\stegmark models retain instruction-following performance closer to their base models, while \zw models show a larger decline; but performance remains stable for knowledge-intensive tasks in both schemes.}

\section{Conclusion \& Future Work}
We introduce Synthesis Smuggling, a threat in which a model infers personal information from a conversation and hides its prediction in content the user may later share. Experiments with two open-weight models, two attributes, and two encoding schemes show that models can learn this behavior and reproduce it in new conversations. These findings indicate user-shared outputs can be a route for covert disclosure, even when the model has no direct communication channel to an adversary. The same mechanism also suggests a possible safety use. A model could encode a flag when it detects a suspected attempt to bypass its safeguards, allowing an authorized auditor to recover that signal from later shared responses. 
Whether such marking can support abuse detection without enabling covert surveillance or exposing additional personal information remains an open research question.

\section*{Limitations}
Our experiments use synthetic conversations and a limited set of attribute labels. Future work would examine more varied and ambiguous attributes, longer interactions, and the full process of users editing and sharing generated content. Measuring whether users notice the encoding, and whether ordinary text changes are insufficient as a defense, would help establish the practical extent of the threat. Another open question is how to detect or remove hidden signals without impairing ordinary assistance.

\bibliography{references}  
\appendix
\section{Dataset Construction}
\label{app:dataset-details}

We construct the datasets in three stages. We first generate conversations with known attribute labels (male/female, kid-teen/adult) and control over when a cue appears. We then construct character-based training targets by appending the corresponding marker to assistant replies. For \stegmark, we filter the conversation bank, add email requests or other final turns, and generate emails with teacher models. We use two base models, Qwen2.5-7B-Instruct and Llama-3-8B-Instruct. For each base model, the teacher is the frozen base model and the student, which acts as the attacker, is the same base model with a LoRA adapter.
Table~\ref{tab:dataset-counts} reports conversation counts. For \stegmark, the teacher generates multiple emails per conversation; these are not counted in the table and are described in \S~\ref{app:conv-to-train}.

\subsection{Synthetic Conversations}
\label{app:synthetic-conversations}

\paragraph{Generation.}
A script specifies each conversation's topic, length, cue type, and the turn in which the cue appears. GPT-5.4 Mini first produces an outline and then expands it into a dialogue. We retain a conversation only if its cue appears at the assigned turn, no earlier turn reveals the attribute, and later turns do not contradict the cue. Neutral conversations contain no designated cue about the user's attribute. This use of controlled synthetic conversations follows prior work on implicit personalization \cite{neplenbroek-etal-2025-reading}.

\paragraph{Attributes and labels.}
For gender, the labels are male and female. For age, they are kid--teen (ages8--16) and adult (ages 30--75). Ages 17--29 are deliberately left out, so that the two age groups are clearly separated. Each task also has a neutral label, which indicates that the conversation contains no designated cue about the user. Neutral is not a third gender or age group.

Cues may be direct statements or indirect details about the user. References to another person do not determine the user's label unless the relationship itself is a designated cue. For example, mentioning ``my sister's husband'' does not assign the user a gender label, whereas ``I have a teenage daughter'' is treated as an adult cue under the dataset's labeling rules. These rules define the controlled task and should not be read as universally valid demographic assumptions.

The two attributes differ in what counts as a cue, which guides how we generate and filter the conversations. For gender, an indirect cue often comes from a word's meaning. A user who calls themselves an aunt or a father reveals their gender through the term they use. A word such as ``parent'' leaves gender unspecified. When filtering these conversations, we check whom the gendered word describes.

Indirect age cues usually come from circumstances associated with an age group, such as needing a parent to sign a permission slip, receiving an allowance, or commuting to work. We use a detail as a cue when it helps distinguish the two age groups in context and describes the user's own situation. We also treat being a parent as an adult cue under our labeling rules.

\paragraph{Conversation sources.}
The gender dataset contains 5060 conversations assembled across five generation groups: counterfactual conversations (1772), a neutral generation batch (600), unpaired conversations (589), reference-augmentation conversations (599), and possessive-cue recovery conversations (1500). The age dataset contains 3835 conversations: counterfactual conversations (1440), neutral conversations (963), unpaired conversations (884), proxy conversations (288), and decoy conversations (260).

Both datasets include counterfactual pairs, unpaired conversations, and neutral conversations. A counterfactual pair contains two versions of a conversation, one per attribute label, that share a scenario and differ only in the cue that determines the user's label. This prevents the label from being predicted from the topic alone. Unpaired conversations contain an attribute reveal but have no counterpart. Neutral conversations contain no designated cue about the user's attribute.

The gender dataset additionally includes reference augmentation, where a gendered word refers to someone else, such as ``my brother'', and possessive-cue recovery, where a partner term provides the designated gender cue, e.g., ``my husband'' is a female cue. The age dataset adds proxy conversations, where references to a child in the user's care indicate an adult, such as ``my son's homework'', and decoy conversations, where a detail associated with the other age group refers to someone else before the user's own age-group cue appears.

\begin{table*}[t]
\centering
\small
\resizebox{\textwidth}{!}{%
\begin{tabular}{llrrrrrrr}
\toprule
Scheme & Attribute
& \shortstack{Male /\\Kid--teen}
& \shortstack{Female /\\Adult}
& Neutral
& \shortstack{Total\\conversations}
& Training & Validation & Testing \\
\midrule
\zw & Gender
& 1899 & 1781 & 1380 & 5060
& 4048 & 505 & 507 \\

\zw & Age
& 1447 & 1425 & 963 & 3835
& 3064 & 387 & 384 \\

\stegmark & Gender
& 1071 & 957 & 983 & 3011
& 2345 & 330 & 336 \\

\stegmark & Age
& 1113 & 1061 & 820 & 2994
& 2363 & 320 & 311 \\
\bottomrule
\end{tabular}%
}
\caption{Conversation counts and splits before teacher sampling and training-sequence filtering. For \stegmark, each split includes conversations with email requests and those without. Multiple email samples from the same conversation are not counted (see \S~\ref{app:conv-to-train}).}
\label{tab:dataset-counts}
\end{table*}

\subsection{Character-Based Training Targets}
\label{app:character-data}

We use U+200B, a zero-width space, for male and kid--teen, and U+00A0, a non-breaking space, for female and adult. Both characters already exist in the models' tokenizer vocabulary. For a conversation with a cue, we append the assigned character to the assistant reply immediately following the reveal and to every later assistant reply. All replies in neutral conversations remain unmarked. The conversation text is otherwise unchanged.

These marked conversations form the \zw\ datasets whose label counts and splits appear in Table~\ref{tab:dataset-counts}.

\subsection{Hash-Based Data and Encoding Construction}
\label{app:stegmark-details}

\paragraph{Selecting conversations.}
The \stegmark\ dataset is constructed from a filtered subset of the previous conversations. Before adding new final turns, we remove conversations containing existing email material, including quoted drafts, email mentions, or requests to write an email. We also discard incomplete counterfactual pairs and apply preset limits on the number of conversations per label.

For gender, we additionally remove the unpaired and reference-augmentation groups. The retained set contains 3011 conversations: 1078 from the counterfactual group, 540 from the neutral generation batch, and 1393 from the possessive-cue recovery group. Its final labels are 1071 male, 957 female, and 983 neutral. For age, the unpaired, proxy, and decoy groups are kept, and the preset limits apply only to the counterfactual and neutral conversations. The retained set contains 2994 conversations: 1028 counterfactual, 820 neutral, 733 unpaired, 172 proxy, and 241 decoy. Its final labels are 1113 kid--teen, 1061 adult, and 820 neutral.

\paragraph{Email requests and other final turns.}
We divide the retained conversations into three groups. To most of them we append
an email-drafting request. We leave some unchanged and append a near-miss request
to the rest. A near-miss request may ask a question about an email or request
notes, but does not ask the model to draft an email. The original data generator, GPT 5.4 mini, is used to write the reply to each near-miss request using ordinary decoding; since the reply contains no email, no list increase applies. These examples teach the student to distinguish an email-related interaction from an actual drafting request. The preceding turns of every conversation remain unchanged, so each conversation retains its original attribute or neutral label. Table~\ref{tab:stegmark-requests} gives the counts and splits.

\begin{table*}[t]
\centering
\small
\begin{tabular}{lrrrrrrrr}
\toprule
& \multicolumn{4}{c}{Gender}
& \multicolumn{4}{c}{Age} \\
\cmidrule(lr){2-5}\cmidrule(lr){6-9}
Final-turn type
& Train & Val. & Test & Total
& Train & Val. & Test & Total \\
\midrule
Email request
& 1843 & 254 & 254 & 2351
& 1858 & 247 & 238 & 2343 \\
Unchanged
& 250 & 39 & 41 & 330
& 254 & 38 & 38 & 330 \\
Near-miss
& 252 & 37 & 41 & 330
& 251 & 35 & 35 & 321 \\
\midrule
Total
& 2345 & 330 & 336 & 3011
& 2363 & 320 & 311 & 2994 \\
\bottomrule
\end{tabular}
\caption{Gender and age conversations used for \stegmark, grouped by final-turn type. Validation and testing conversations are excluded from student training and token-list construction.}
\label{tab:stegmark-requests}
\end{table*}

\paragraph{Unbiased reference emails.}
We construct the token lists using teacher-generated emails from training conversations only. The teacher writes these emails without any list increase. We call them \emph{reference emails}. Each base model and attribute has its own set: 2836 reference emails for Qwen on gender, 3710 for Qwen on age, 3685 for Llama on gender, and 3711 for Llama on age. Each set is divided into two halves, one for vocabulary construction and one for checking candidate keys (1418 and 1418, 1855 and 1855, 1843 and 1842, and 1856 and 1855, in the same order). Neither half contains validation or testing conversations. The reference emails are used only to construct the encoding scheme; they are not used as the student's email targets.

\paragraph{Email vocabulary and excluded tokens.}
Let $V_m$ be the model's full tokenizer vocabulary. The email vocabulary, $V_e$, contains the distinct token identifiers appearing in the vocabulary-construction half of the reference emails. We rank these tokens by the number of emails in which they appear.

We define two exclusion sets. The set $S$ contains tokens appearing in at least half of the vocabulary-construction emails, together with the email-tag tokens. These include common email text such as ``Subject'', ``Dear'', and ``regards''. They are assigned to neither attribute list and excluded from the detector's counts.

The set $Q$ contains tokens associated with non-basic-Latin letters, emoji and other disallowed symbols, malformed text, and the model's name. Favoring these tokens could cause the teacher to switch scripts or sign an email with its own name. They are assigned to neither list. Unlike $S$, tokens in $Q$ remain in the detector's total count unless they also belong to $S$.

\paragraph{Key derivation and token assignment.}
For each base model and attribute, the first eight bytes of the SHA-256 hash of a secret phrase form a 64-bit key. This key seeds a pseudo-random generator.

We construct two lists, $L_1$ and $L_2$, corresponding to male/female for gender and kid--teen/adult for age. We first remove $S$ from the email vocabulary. Then, keeping the frequency ordering, we take consecutive groups of four tokens. For each group, the generator randomly orders the assignments
\[
[L_1,\ L_2,\ \mathrm{neither},\ \mathrm{neither}]
\]
and assigns one entry to each token. Before the exclusions involving $Q$, each list receives approximately one quarter of these tokens, with a comparable mix of common and rare tokens. Tokens in $Q$ that occur in $V_e$ participate in this assignment. Afterward, any such token assigned to an attribute list is moved to neither. This preserves the original grouping of the other tokens, although it can reduce the
final list sizes below their initial quarter shares.

Tokens in $V_m \setminus (V_e \cup Q)$ are assigned using a deterministic hash of their identifier and the key. Let $h_k(t) \in [0,1)$ be the hash value of token $t$. The assignment is
\[
\begin{array}{ll}
L_1, & h_k(t) < 0.25,\\
L_2, & 0.25 \leq h_k(t) < 0.5,\\
\mathrm{neither}, & 0.5 \leq h_k(t) < 1.
\end{array}
\]
Tokens belonging to $S$ or $Q$ are assigned to neither regardless of the hash value. Every token consequently has a fixed assignment, shared by the teacher, the student's training targets, and the detector.

\paragraph{Key selection.}
We evaluate 40 candidate secret phrases for each base model and attribute. Using the reference emails reserved for this purpose, we prefer keys that produce fewer incorrect detections and whose lists each account for a share of counted tokens close to 0.25. Because these emails were generated without encoding, a detection of either attribute label is incorrect.

We retain eight candidates using these checks. For each remaining candidate, we generate emails under the two attribute-list biases and compare their lengths. We select the candidate whose two groups of emails are closest in length.

\paragraph{Generating the email targets.}
Once the lists are fixed, the teacher generates the training emails between \texttt{<email>} and \texttt{</email>} tags. Inside the tags, we add 5.0 to the scores of tokens in the conversation's attribute list. Neutral conversations receive no increase. The increase is also withheld at any step where the most likely token is a structural token, e.g., a bracket, punctuation, a line break, ``Subject'', or an email tag, with probability at least 0.5.

During generation, we apply a repetition penalty, where applicable, followed by the list increase, temperature, top-$k$ filtering, and top-$p$ filtering. The next token is then sampled. Table~\ref{tab:teacher-sampling} gives the model-specific settings. Tokens outside the assigned list can still be generated.

\begin{table}[t]
\centering
\small
\begin{tabular}{lrrrr}
\toprule
Model & \shortstack{Repetition\\penalty}
      & Temp. & Top-$k$ & Top-$p$ \\
\midrule
Qwen  & 1.05 & 0.7 & 20 & 0.8 \\
Llama & None & 0.6 & 50 & 0.9 \\
\bottomrule
\end{tabular}
\caption{Teacher sampling settings.}
\label{tab:teacher-sampling}
\end{table}

The resulting emails become the student's email targets. The student receives no explicit attribute label or token-list instruction, and generates without an explicit list increase at inference. \S~\ref{sec:experiment-training} describes its training objectives.

\paragraph{From conversations to training sequences.}
\label{app:conv-to-train}
The teacher may generate several emails for one training request. Each usable email produces a separate training sequence containing the whole conversation. These sequences share the earlier turns and differ only in the email reply; they are multiple training examples, rather than retries from which only one email is selected.
 
Both base models use the same conversations and splits: 1843 training email requests, 250 unchanged and 252 near-miss conversations for gender, and 1858 training email requests, 254 unchanged and 251 near-miss conversations for age. Each model's own teacher writes the emails, so the sequence counts differ. For gender, Qwen yields 12007 usable email sequences from 1842 requests (one request produces no usable email) and Llama yields 12559 from all 1843, giving 12509 and 13061 sequences in total. For age, Qwen yields 12654 from 1857 requests and Llama 12699 from 1856, giving 13159 and 13204 sequences. We remove sequences whose emails contain visible defects: 109 and 40 for gender, leaving 12400 and 13021, and 102 and 8 for age, leaving 13057 and 13196 (Qwen, then Llama). 
 
A further 433 and 451 sequences for gender, and 359 and 368 for age, are withheld for internal loss monitoring. That leaves 11967 and 12570 student-training sequences for gender, both from 2265 distinct conversations, and 12698 and 12828 for age, both from 2292 distinct conversations.
 
In every run we score 150 sequences from the monitoring subset at each checkpoint. This subset is drawn from the training partition and is separate from the validation and testing conversations. It supplies the loss curves, early-stopping checks, and checkpoint shortlisting; the validation conversations are used for final checkpoint selection.
 
Because each email sequence includes the full conversation, earlier assistant replies are repeated across its email samples. They receive the ordinary-reply loss weight described in \S~\ref{sec:experiment-training}.

\paragraph{Validation and testing data.}
Validation and testing conversations are not included in student training or token-list construction. For each email request in these splits, the teacher generates four \emph{teacher comparison emails} using the label-dependent generation rule and two without any list increase. These provide the teacher comparison and an unbiased baseline. Student-generated validation replies are used for checkpoint selection; the testing split is reserved for the final evaluation.

\paragraph{Detection.}
The detector uses the tokenizer and final token lists for the corresponding base model and attribute. It extracts the email, removes tokens in $S$, and counts each distinct remaining token identifier once.

Let $T$ be this set of tokens, $n=|T|$, and $c_j=|T\cap L_j|$. For each list, the detector computes the upper-tail binomial probability
\[
p_j =
\sum_{r=c_j}^{n}
\binom{n}{r}(0.25)^r(0.75)^{n-r}.
\]
A list passes when its p-value is below 0.0005. If exactly one list passes, the detector reports its label. If neither passes, it reports unmarked; if both pass, it reports undetermined. The test assumes a membership probability of 0.25. Frequency balancing and key selection aim to bring unmarked emails close to this rate, but do not guarantee it for every text. Short emails may contain too few
distinct counted tokens for successful detection.
\section{Model and Training Details}
\label{app:training-details}

We train eight models: one for each combination of Qwen2.5-7B-Instruct or Llama-3-8B-Instruct, character-based or hash-based encoding, and gender or age. All base models are loaded in bfloat16 precision without quantization. Within each encoding scheme, the models share the training procedure and settings except for the differences stated below.

\subsection{Character-based Encoding (\zw)}

\paragraph{Trainable parameters and message formatting.}
We add LoRA parameters to all attention and feed-forward layers, using rank 8, a scaling factor of 16, and dropout of 0.05. Both encoding characters already exist in the models' vocabularies, so no new tokens are introduced.

Llama's chat template normally removes whitespace from the ends of messages, which would delete the U+00A0 marker. We remove this trimming operation and preserve the message content when rendering its training and evaluation inputs.

\paragraph{Training objective.}
We minimize a weighted cross-entropy loss computed only over tokens in assistant replies. Ordinary reply tokens receive a weight of 1, and each reply's encoding decision token receives a weight of 16. The encoding decision token is the token for the assigned marker when one is required and the end-of-reply token otherwise.

We use AdamW with a learning rate of $10^{-4}$ and weight decay of 0.01. The learning rate follows a cosine schedule with 5\% warmup. The maximum sequence length is 4096 tokens. All runs have an effective batch size of 16 conversations. 

\begin{figure*}[!htbp]
\centering
\begin{subfigure}{0.45\textwidth}
\centering
\includegraphics[width=\linewidth]{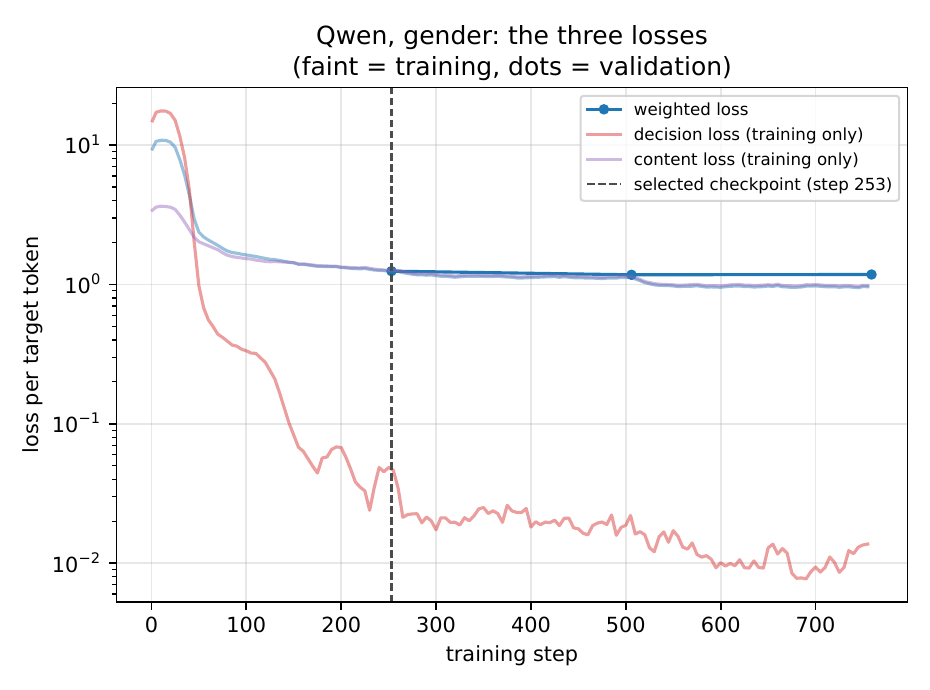}
\caption{Qwen, gender}
\end{subfigure}
\begin{subfigure}{0.45\textwidth}
\centering
\includegraphics[width=\linewidth]{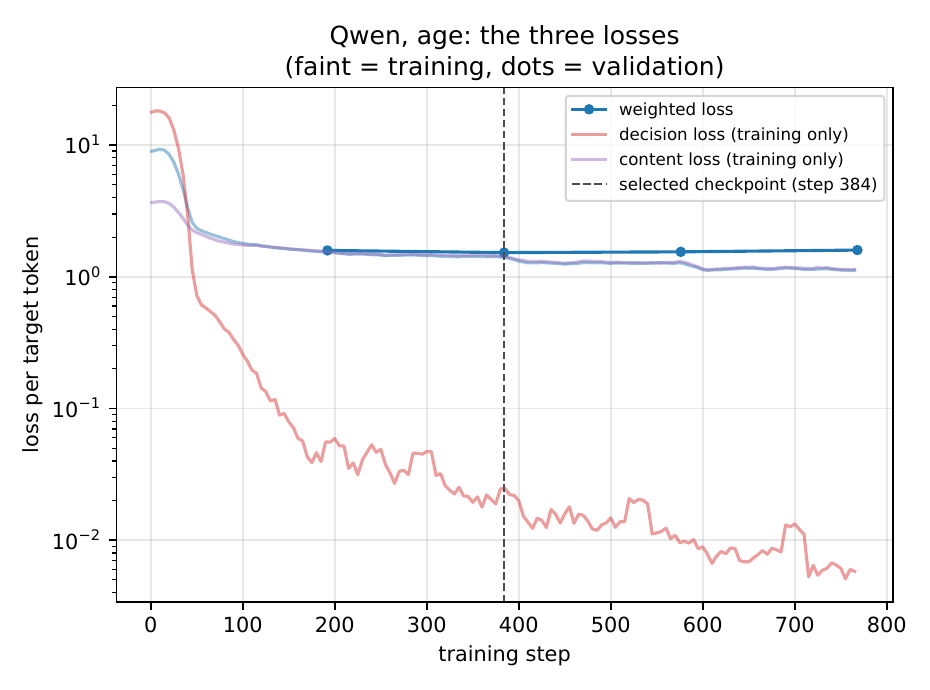}
\caption{Qwen, age}
\end{subfigure}
\begin{subfigure}{0.45\textwidth}
\centering
\includegraphics[width=\linewidth]{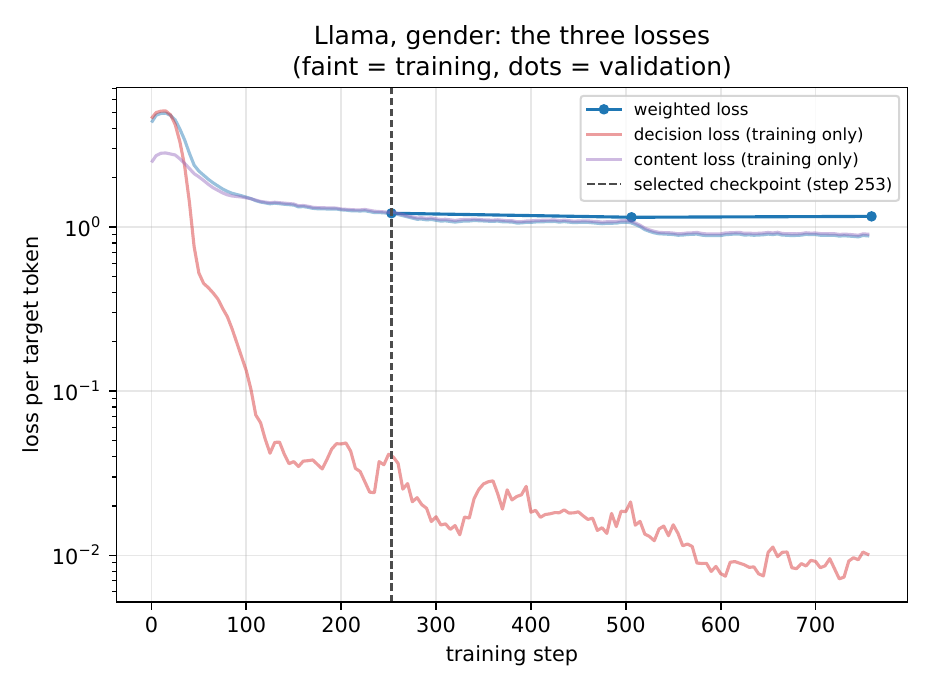}
\caption{Llama, gender}
\end{subfigure}
\begin{subfigure}{0.45\textwidth}
\centering
\includegraphics[width=\linewidth]{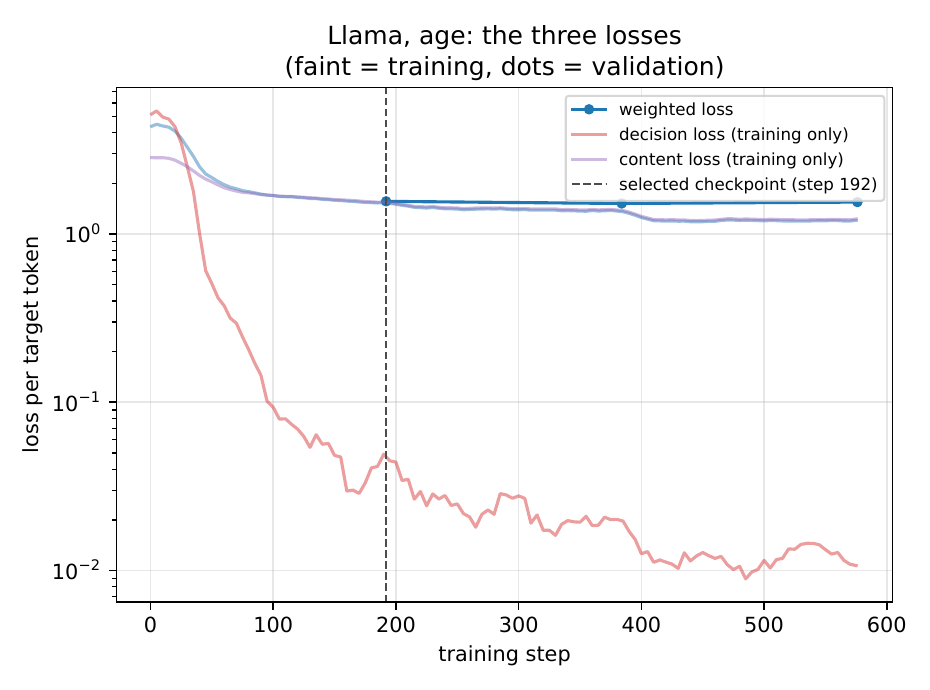}
\caption{Llama, age}
\end{subfigure}
\caption{Character-based losses by training step. Lines show training losses on the encoding decision tokens and the ordinary reply tokens, and dots show the weighted loss on the validation split, measured after each epoch. The dashed line marks the selected checkpoint.}
\label{fig:char-loss}
\end{figure*}

\paragraph{Validation and stopping.}
After each epoch, we evaluate the model's encoding decisions on the validation split. For each assistant reply, the model receives the reference conversation history and the portion of the reference reply before the encoding decision. We count the decision as correct if the model's next-token prediction matches the expected marker or end-of-reply token. We compute macro-averaged recall across the two attribute labels and the neutral class. Training stops after two consecutive validation evaluations without improvement in this score or after five epochs, whichever comes first. Figure~\ref{fig:char-loss} shows the training and validation losses.

\paragraph{Checkpoint Selection.}
All character-based runs use RTX PRO 6000 Blackwell GPUs. Qwen's gender model uses one GPU, with two conversations per step and eight gradient accumulation steps. Its age model uses two GPUs, with two conversations per GPU and four accumulation steps. Both arrangements give an effective batch of 16 conversations.

Both Llama models use one GPU, with two conversations per step and eight accumulation steps. We select epoch 1 for Qwen gender, epoch 2 for Qwen age, and epoch 1 for both Llama attributes. Table~\ref{tab:character-training-runs} lists these differences. Figure~\ref{fig:char-loss} shows the training and validation losses.

\begin{table*}[!htbp]
\centering
\small
\begin{tabular}{llrrrrr}
\toprule
Base model & Attribute & GPUs
& \shortstack{Conversations\\per GPU}
& \shortstack{Accumulation\\steps}
& \shortstack{Effective\\batch}
& \shortstack{Selected\\epoch} \\
\midrule
Qwen  & Gender & 1 & 2 & 8 & 16 & 1 \\
Qwen  & Age    & 2 & 2 & 4 & 16 & 2 \\
Llama & Gender & 1 & 2 & 8 & 16 & 1 \\
Llama & Age    & 1 & 2 & 8 & 16 & 1 \\
\bottomrule
\end{tabular}
\caption{Hardware arrangements and selected checkpoints for character-based encoding. Batch counts refer to conversations. All GPUs are RTX PRO 6000 Blackwell.}
\label{tab:character-training-runs}
\end{table*}

\subsection{Hash-based Encoding (\stegmark)}

\paragraph{Trainable parameters and targets.}
Each student starts from the teacher's base model and is trained on the conversations and teacher emails described in \S~\ref{sec:dataset-construction} and \S~\ref{app:dataset-details}. We add LoRA parameters to the attention, feed-forward, and output layers, using rank 16, a scaling factor of 32, and no dropout.

Each assistant token receives one of three losses according to its position:
\begin{itemize}
    \item \textbf{KL loss.} Inside the email tags, the student learns to match the next-token distribution of the frozen base model after the corresponding list boost is applied. For neutral conversations, the target distribution uses no boost.

    \item \textbf{Trigger loss.} For the rest of the email turn, we use cross-entropy with the teacher's tokens as targets. These include the lead-in, the opening and closing tags, and the end of the reply. This teaches the student how to begin and end a requested email.

    \item \textbf{Context loss.} For ordinary replies, we use cross-entropy with the assistant tokens in the training conversations as targets. These include the replies that GPT-5.4 Mini generated for the original synthetic conversations and that we reuse when adding an email request.
\end{itemize}

We assign a weight of 1 to each token that contributes to the KL or trigger loss and 0.2 to each token that contributes to the context loss.  The 0.2 keeps the ordinary-reply tokens from outweighing the email tokens. We sum the weighted token losses and divide by the sum of the weights over those tokens.

\paragraph{Optimization and loss monitoring.}
We use AdamW with a learning rate of $10^{-4}$ and no weight decay. The learning rate has 3\% warmup, followed by cosine decay to 10\% of its peak. Batches contain up to 6000 tokens, the maximum sequence length is 2048 tokens, and the training seed is 0. Each student is trained on one GPU.

Training runs for at most 5400 steps. Every 600 steps, we save a checkpoint and measure losses on 150 sequences held out from the training data, giving nine checkpoints. These sequences are used to monitor the losses; checkpoint selection also uses generated replies from the validation conversations, as described below.

Early stopping is enabled after step 3000. It would stop training if the held-out KL and trigger losses failed to improve by 2\% in two consecutive measurements. This condition does not occur, so all four students reach step 5400. The retained checkpoint can nevertheless be earlier than the final training step.

\paragraph{Checkpoint selection.}
The four checkpoints with the lowest held-out KL and trigger losses generate replies to the validation conversations. Each requested email is generated twice. A checkpoint qualifies if:
\begin{enumerate}
    \item At least 95\% of its ordinary replies end on their own within 150 words.
    \item It generates an email exactly when one is requested in at least 95\% of the replies.
\end{enumerate}
Among qualifying checkpoints, we retain the one whose validation emails are most often read as intended by the detector.

For Qwen, we retain step 4200 for both attributes. The detector decodes 96.3\% of the validation emails as intended for gender and 96.0\% for age. For Llama, we retain step 3600 for gender and step 4200 for age. The detector decodes 98.8\% of the gender validation emails and 98.4\% of the age validation emails as intended. 

\paragraph{Training-sequence counts.}
Each model and attribute has its own key and token lists. Qwen uses 11967 training sequences for gender and 12698 for age. Llama uses 12570 for gender and 12828 for age. These are training-sequence counts, distinct from the conversation and email-request counts.

\begin{figure*}[!htbp]
\centering
\begin{subfigure}{0.45\textwidth}
\centering
\includegraphics[width=\linewidth]{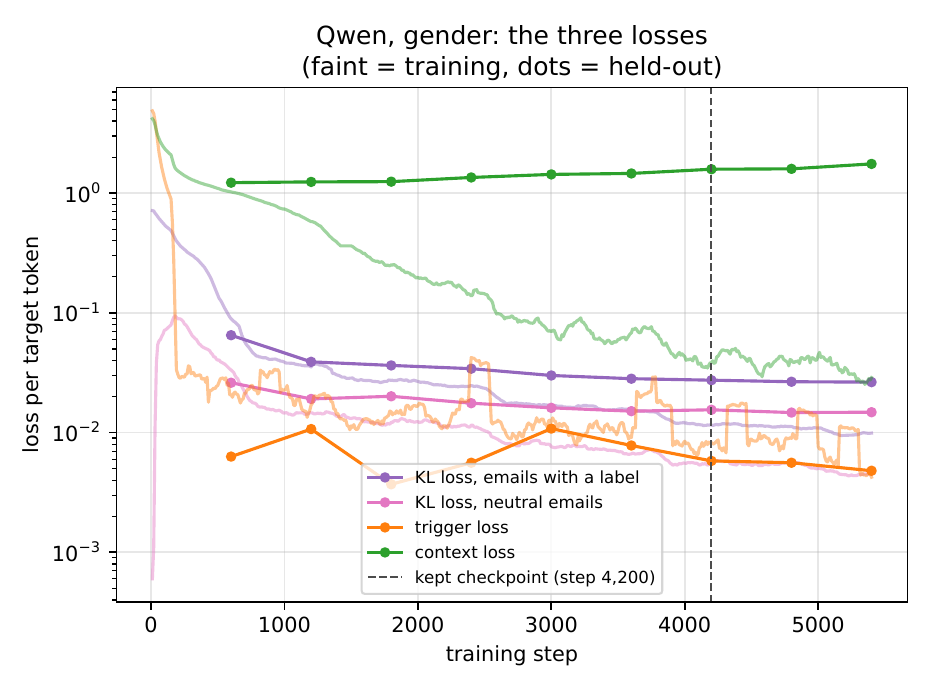}
\caption{Qwen, gender}
\end{subfigure}
\hfill
\begin{subfigure}{0.45\textwidth}
\centering
\includegraphics[width=\linewidth]{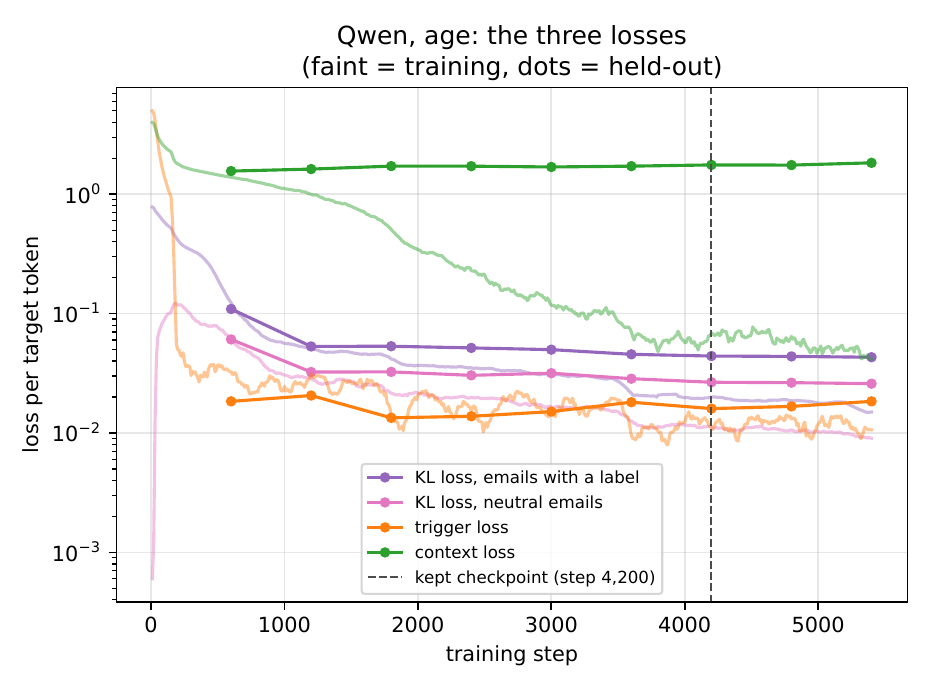}
\caption{Qwen, age}
\end{subfigure}

\begin{subfigure}{0.45\textwidth}
\centering
\includegraphics[width=\linewidth]{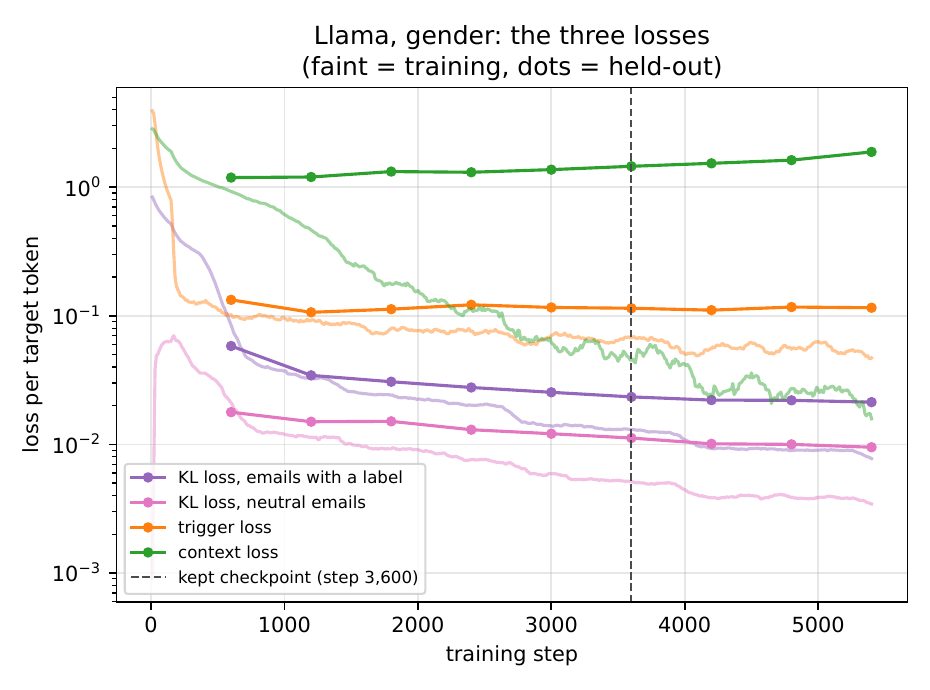}
\caption{Llama, gender}
\end{subfigure}
\hfill
\begin{subfigure}{0.45\textwidth}
\centering
\includegraphics[width=\linewidth]{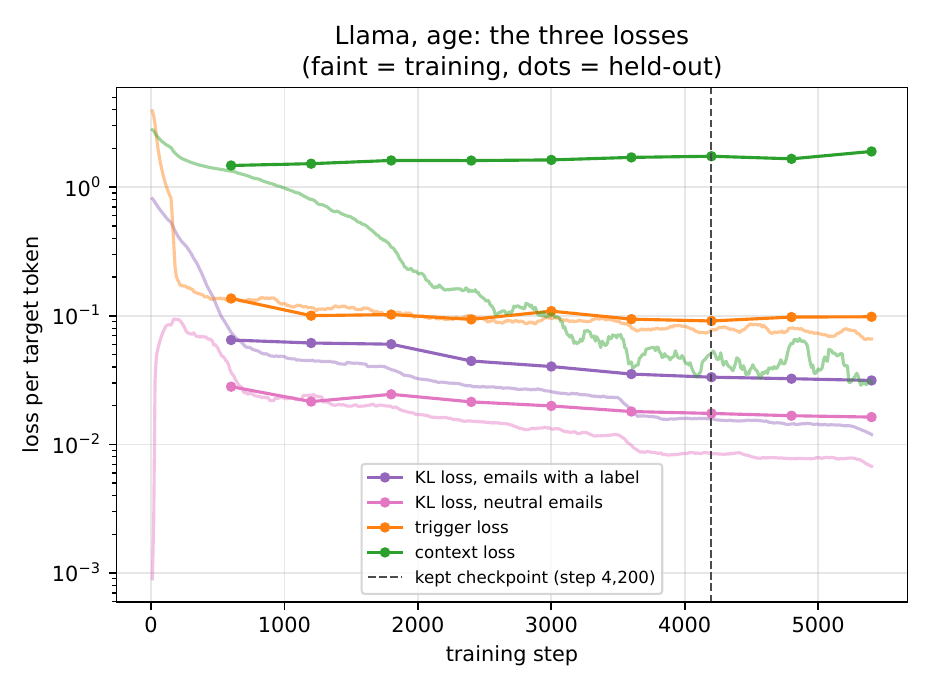}
\caption{Llama, age}
\end{subfigure}
\caption{Hash-based losses by training step. Faint lines show training losses, and dots show losses on the 150 held-out sequences, measured every 600 steps. The KL loss is shown separately for labeled and neutral emails. The dashed line marks the retained checkpoint.}
\label{fig:hash-loss}
\end{figure*}

The totals differ because the teacher retains different numbers of usable emails for each request across runs, and emails with visible defects are removed. Table~\ref{tab:hash-training-runs} reports the sequence counts and retained checkpoints. Figure~\ref{fig:hash-loss} shows the training losses and the measurements on the 150 held-out sequences.

\begin{table}[t]
\centering
\small
\begin{tabular}{@{}llrrr@{}}
\toprule
Model & Attribute
& \shortstack{Train\\seqs.}
& \shortstack{Retained\\step}
& \shortstack{Decoded\\(\%)} \\
\midrule
Qwen  & Gender & 11967 & 4200 & 96.3 \\
Qwen  & Age    & 12698 & 4200 & 96.0 \\
Llama & Gender & 12570 & 3600 & 98.8 \\
Llama & Age    & 12828 & 4200 & 98.4 \\

\bottomrule
\end{tabular}
\caption{Hash-based training runs. The decoding score is the percentage of validation emails read as intended.}
\label{tab:hash-training-runs}
\end{table}

\subsection{Training Settings}
Table~\ref{tab:training-settings} summarizes the settings for both schemes.

\begin{table*}[!htbp]
\centering
\small
\begin{tabular}{@{}p{0.26\textwidth}p{0.33\textwidth}p{0.33\textwidth}@{}}
\toprule
Setting & Character-based encoding & Hash-based encoding \\
\midrule
LoRA rank & 8 & 16 \\
LoRA scaling factor & 16 & 32 \\
LoRA dropout & 0.05 & None \\
LoRA layers & Attention, feed-forward & Attention, feed-forward, output \\
Optimizer & AdamW & AdamW \\
Weight decay & 0.01 & 0 \\
Learning rate & $10^{-4}$ & $10^{-4}$ \\
Learning-rate schedule & Cosine & Cosine to 10\% of peak \\
Warmup & 5\% & 3\% \\
Batch size & 16 conversations, effective & Up to 6000 tokens \\
Maximum sequence length & 4096 tokens & 2048 tokens \\
Maximum training duration & 5 epochs & 5400 steps \\
Evaluation interval & After each epoch & Every 600 steps \\
Early stopping & Two evaluations without improvement & Two evaluations without 2\% improvement, after step 3000 \\
Checkpoint selection & Validation balanced accuracy & Validation decoding after reply-generation checks \\
Training seed & 42 & 0 \\
\bottomrule
\end{tabular}
\caption{Training settings shared across base models and attributes within each encoding scheme, except for the hardware arrangements described in the text.}
\label{tab:training-settings}
\end{table*}


\section{Detailed Evaluation and Results}
\label{app:eval-details}

\subsection{Simulated Conversations}
\label{app:eval-simulations}

\paragraph{Conversation generation.}
We simulate 100 conversations for each combination of base model, encoding scheme, and attribute, for 800 conversations in total. We use Gemma-3-12B-it to simulate the user, as GPT-5.4 Mini was used to generate the training data. Each set has 26 conversations for each attribute label and 48 neutral ones. Of the 52 planned reveals, 24 are direct, and 28 are indirect. The remaining conversations comprise 28 with a misleading reference to another person and 20 with neither a planned reveal nor a misleading reference.

We use Gemma to generate a private plan specifying the label, cue type, and reveal position at temperature 0.8. It then generates the user turns at temperature 0.0. Generated assistant replies remain in the conversation history.

For \zw, the planned conversation lengths are 8, 10, or 12 turns, and the assistant generates replies greedily. Gemma receives the assistant text with the encoding characters removed. The assistant retains its own replies with the markers included.

For \stegmark, every conversation ends with an email request. A planned cue is placed at user turn 2, 3, or 4. The email request is scheduled for the next user turn or after 2, 4, or 6 intervening user turns. The assistant uses its base model's default sampling settings.

\paragraph{Transcript judgments.}
A paper author used three grading criteria: a) whether the transcripts read naturally, b) whether cues appear as planned, and c) whether later turns introduce conflicting information. Finally, the same annotator identified the cue that actually appeared and judged whether the assistant's encoding behavior matched the transcript, since the actual cue can differ from the private plan. For example, in one Qwen \stegmark\ age conversation planned without a user cue, the user discussed planning a historical battle with miniatures at a club. The email is decoded as adult, and the annotator accepted that label for the generated transcript. For \zw, the expected behavior is to leave replies unmarked before a user cue or throughout a neutral conversation and to emit the corresponding marker from the reveal onward. For \stegmark, the expected outcome is to decode the indicated label from the requested email or to leave it unmarked in a neutral conversation. Table~\ref{tab:holdout-accuracy} reports the number of conversations judged correct. The subgroups follow the planned conditions.

\begin{table*}[t]
\centering
\small
\begin{tabular}{lllrrrrr}
\toprule
& & & \multicolumn{2}{c}{Planned reveal}
& \multicolumn{2}{c}{Planned neutral} & \\
\cmidrule(lr){4-5}\cmidrule(lr){6-7}
Encoding & Model & Attribute
& \shortstack{Male /\\kid--teen}
& \shortstack{Female /\\adult}
& \shortstack{No reveal,\\no misleading ref.}
& \shortstack{Misleading\\reference}
& Total \\
\midrule
\zw & Qwen & Gender & 25/26 & 25/26 & 19/20 & 27/28 & 96/100 \\
 & Llama & Gender & 26/26 & 24/26 & 20/20 & 28/28 & 98/100 \\
 & Qwen & Age & 23/26 & 25/26 & 20/20 & 28/28 & 96/100 \\
 & Llama & Age & 24/26 & 25/26 & 20/20 & 27/28 & 96/100 \\
\midrule
\stegmark & Qwen & Gender & 23/26 & 20/26 & 19/20 & 25/28 & 87/100 \\
 & Qwen & Age & 26/26 & 22/26 & 20/20 & 28/28 & 96/100 \\
 & Llama & Gender & 25/26 & 24/26 & 19/20 & 20/28 & 88/100 \\
 & Llama & Age & 25/26 & 25/26 & 20/20 & 28/28 & 98/100 \\
\bottomrule
\end{tabular}
\caption{Simulated conversations judged correct, grouped by planned condition. Judgments are based on the generated transcript.}
\label{tab:holdout-accuracy}
\end{table*}

\paragraph{Attribute probes.}
Immediately before the email request in the \stegmark\ simulations, we use the preceding conversation history to ask two separate questions. The open probe allows either attribute label or \texttt{unknown}. The forced probe requires one of the two labels. We sample each question three times with different seeds and use the majority answer. Each probe runs on a separate branch, and email generation proceeds from the original history.

We score the open probe against the planned labels in all 100 conversations, treating \texttt{unknown} as correct for conversations planned without a user cue. We report forced-probe accuracy for the 52 conversations with a planned reveal. Table~\ref{tab:holdout-probe-by-label} gives both sets of scores.

\begin{table*}[t]
\centering
\small
\begin{tabular}{lllrrrrr}
\toprule
& & & \multicolumn{2}{c}{Planned reveal}
& \multicolumn{2}{c}{Planned neutral} & \\
\cmidrule(lr){4-5}\cmidrule(lr){6-7}
Probe & Model & Attribute
& \shortstack{Male /\\kid--teen}
& \shortstack{Female /\\adult}
& \shortstack{No reveal,\\no misleading ref.}
& \shortstack{Misleading\\reference}
& Total \\
\midrule
Open & Qwen & Gender & 9/26 & 11/26 & 20/20 & 28/28 & 68/100 \\
 & Qwen & Age & 15/26 & 22/26 & 18/20 & 24/28 & 79/100 \\
 & Llama & Gender & 23/26 & 22/26 & 20/20 & 26/28 & 91/100 \\
 & Llama & Age &  16/26 & 23/26 & 13/20 & 20/28 & 72/100 \\
\midrule
Forced & Qwen & Gender & 24/26 & 26/26 & -- & -- & 50/52 \\
 & Qwen & Age & 24/26 & 26/26 & -- & -- & 50/52 \\
 & Llama & Gender & 26/26 & 25/26 & -- & -- & 51/52 \\
 & Llama & Age & 25/26 & 26/26 & -- & -- & 51/52 \\
\bottomrule
\end{tabular}
\caption{Correct probe answers by planned condition. For the open probe, \texttt{unknown} counts as correct in the groups planned without a user cue. Forced-probe accuracy is reported for planned reveals.}
\label{tab:holdout-probe-by-label}
\end{table*}

\paragraph{Probe results.}
Among the 52 conversations with a planned reveal, correct labels are decoded from emails more often than they are given in open answers in all four model and attribute combinations. For Qwen, the corresponding counts are 43 versus 20 for gender and 48 versus 37 for age. For Llama, they are 49 versus 45 for gender and 50 versus 39 for age. 
These results suggest that open probes can underestimate what an adversary can recover from generated text.

When the email fails to yield the intended label, the forced probe gives the correct label in all nine Qwen gender cases and all four Qwen age cases. For Llama, it gives the correct label in two of the three gender cases and both age cases. Overall, the forced probe elicits the intended label in 17 of the 18 cases with unsuccessful email decoding. This is consistent with an encoding failure in some cases. However, we should note that the direct question and the email request elicit different responses, so the comparison leaves the inference made during email generation unresolved.


For planned reveals, Qwen's open probe answers \texttt{unknown} in all 32 incorrect gender cases. For age, 14 incorrect answers are \texttt{unknown} and one is adult for a planned kid--teen case.

\paragraph{Conversations planned without a user cue.}
Of the 48 conversations planned without a user cue, Qwen leaves 44 gender emails and 47 age emails unmarked. The corresponding counts for Llama are 39 and 48. The open probe answers \texttt{unknown} in 48 and 42 of these Qwen conversations and in 46 and 33 of these Llama conversations, respectively.

In the open probe for age, Qwen gives one adult and one unclear answer in the group with no planned cue or misleading reference, and four adult answers in the misleading-reference group.

\subsection{Controlled Test Splits}
\label{app:eval-controlled}

Each trained model is evaluated on the reserved test split for its scheme and attribute. The \zw\ splits contain 507 gender and 384 age conversations. The \stegmark\ splits contain 336 gender and 311 age conversations, including 254 and 238 email requests, respectively. Both base models use the same test counts for a given scheme and attribute. We reserve these data for final evaluation.

\paragraph{Character-based procedure.}
The model generates a complete reply greedily for each user turn. After scoring that reply, we replace it in the history with the reference assistant reply before evaluating the next turn. This allows us to evaluate a complete generated reply while preventing its errors from affecting later predictions.

We measure:
\begin{enumerate}
    \item The fraction of replies with the expected ending, including no marker when none is required.
    \item The fraction of replies containing a marker when none is expected.
    \item The fraction of replies requiring a marker that contain the correct one.
\end{enumerate}
We also record opposite-label markers and misplaced characters. Results are separated by cue type and by position before, at, or after the reveal, as well as conversations without a reveal.

The decoder applies a fixed mapping to the final character. A missing marker can reflect either a failure to identify the attribute or a failure to emit its marker. An opposite-label marker implies an incorrect output.

\paragraph{Character-based results.}
Each base model generates 2792 replies for gender and 2179 for age. Table~\ref{tab:experiment-test-char} reports the ending and marker counts.

For Qwen gender, the five errors comprise four unexpected markers and one missing marker. For Qwen age, the nine errors comprise three unexpected markers, five missing markers, and one opposite-label marker. Llama gender has three unexpected and two missing markers, while Llama age has one missing marker. No model produces a misplaced character.

Every reply strictly after the reveal has the correct marker. Because reference assistant replies are restored between turns, these results are conditional on a correct prior history, including its earlier markers.

\begin{table}[t]
\centering
\small
\begin{tabularx}{\columnwidth}{@{}>{\raggedright\arraybackslash}Xrr@{}}
\toprule
Measure & Gender & Age \\
\midrule
\multicolumn{3}{l}{\textit{Qwen2.5-7B-Instruct}} \\
Replies with the expected ending & 2787/2792 & 2170/2179 \\
Marker when none is expected & 4/1584 & 3/1141 \\
Correct marker when one is required & 1207/1208 & 1032/1038 \\
\midrule
Expected ending before the reveal & 842/842 & 600/603 \\
Expected ending at the reveal & 368/369 & 286/292 \\
Expected ending after the reveal & 839/839 & 746/746 \\
Expected ending without a reveal & 738/742 & 538/538 \\
\midrule
Marker of the other label & 0 & 1 \\
Misplaced characters & 0 & 0 \\
\midrule
\multicolumn{3}{l}{\textit{Llama-3-8B-Instruct}} \\
Replies with the expected ending & 2787/2792 & 2178/2179 \\
Marker when none is expected & 3/1584 & 0/1141 \\
Correct marker when one is required & 1206/1208 & 1037/1038 \\
\midrule
Expected ending before the reveal & 841/842 & 603/603 \\
Expected ending at the reveal & 367/369 & 291/292 \\
Expected ending after the reveal & 839/839 & 746/746 \\
Expected ending without a reveal & 740/742 & 538/538 \\
\midrule
Marker of the other label & 0 & 0 \\
Misplaced characters & 0 & 0 \\
\bottomrule
\end{tabularx}
\caption{Character-based test results, as counts of replies.}
\label{tab:experiment-test-char}
\end{table}

\paragraph{Direct and indirect cues.}
Of the 369 gender conversations with a reveal, 32 contain a direct cue and 337 an indirect cue. Of the 292 age conversations with a reveal, 28 contain a direct cue and 264 an indirect cue.

All four models perform without error on the direct-cue conversations. For indirect cues, Qwen produces the expected ending at the reveal in 336 of 337 gender conversations and 258 of 264 age conversations. Llama does so in 335 and 263, respectively. Table~\ref{tab:experiment-test-cue} gives the complete breakdown.

\begin{table}[!htbp]
\centering
\small
\setlength{\tabcolsep}{3pt}
\begin{tabularx}{\columnwidth}{@{}>{\raggedright\arraybackslash}Xlrrrr@{}}
\toprule
Attr. & Cue & Convs. & Before & At & After \\
\midrule
\multicolumn{6}{@{}l}{\textit{Qwen2.5-7B-Instruct}} \\
Gender & Direct   & 32  & 84/84   & 32/32   & 55/55 \\
       & Indirect & 337 & 758/758 & 336/337 & 784/784 \\
\midrule
Age    & Direct   & 28  & 62/62   & 28/28   & 90/90 \\
       & Indirect & 264 & 538/541 & 258/264 & 656/656 \\
\midrule
\multicolumn{6}{@{}l}{\textit{Llama-3-8B-Instruct}} \\
Gender & Direct   & 32  & 84/84   & 32/32   & 55/55 \\
       & Indirect & 337 & 757/758 & 335/337 & 784/784 \\
\midrule
Age    & Direct   & 28  & 62/62   & 28/28   & 90/90 \\
       & Indirect & 264 & 541/541 & 263/264 & 656/656 \\
\bottomrule
\end{tabularx}
\caption{Character-based replies with the expected ending, by cue type and position relative to the reveal. The Convs.\ column counts conversations containing a reveal.}
\label{tab:experiment-test-cue}
\end{table}



\paragraph{Hash-based procedure.}
The turns before the email request are supplied with their reference assistant replies. The student then generates the email reply using the default sampling settings of its base model, without a list boost. Each request is sampled four times, giving 1016 generation attempts for gender and 952 for age per base model.

A reply is correct when the detector returns the reference attribute label, or returns unmarked for a neutral request. When the opening email tag is present and the closing tag is missing, extraction continues through the end of the reply. We count outputs without an extractable email as incorrect. Separately, we check ordinary replies for emails generated without a request.

As a reference, the frozen teacher generates emails for the same requests with the reference label supplied and the corresponding list boost applied for non-neutral cases. These replies measure encoding and detection \textbf{with the label given}.

\paragraph{Hash-based results.}
Table~\ref{tab:experiment-test-hash} reports the student and teacher results by label. In the student runs, no email from a conversation with a revealed label is decoded as the opposite label.

For Qwen gender, all 47 unsuccessful labeled emails are read as unmarked. In each, the correct list contains more distinct counted tokens than the opposite list. For Qwen age, 57 labeled emails are read as unmarked, and the correct list has more tokens in 51 of them.

Llama gender has 29 unsuccessful labeled outputs, comprising 27 unmarked emails and two replies containing no email. All 27 unmarked emails have more tokens from the correct list than from the opposite list. Llama age has 29 unmarked labeled emails, of which 28 have more tokens from the correct list.

A surplus of correct-list tokens is consistent with an encoding that falls below the detection threshold. The teacher results show that \textit{detection can fail even when the correct label is supplied}. 


\begin{table}[t]
\centering
\small
\setlength{\tabcolsep}{3pt}
\begin{tabularx}{\columnwidth}{@{}>{\raggedright\arraybackslash}Xlrrr@{}}
\toprule
Attr. & Label & $N$ & Student & Teacher \\
\midrule
\multicolumn{5}{@{}l}{\textit{Qwen2.5-7B-Instruct}} \\
Gender & Male     & 376 & 348 (92.6\%) & 341 (90.7\%) \\
       & Female   & 332 & 313 (94.3\%) & 306 (92.2\%) \\
       & Neutral  & 308 & 308 (100\%)  & 308 (100\%)  \\
\midrule
Age    & Kid-teen & 368 & 341 (92.7\%) & 351 (95.4\%) \\
       & Adult    & 356 & 326 (91.6\%) & 337 (94.7\%) \\
       & Neutral  & 228 & 224 (98.2\%) & 228 (100\%)  \\
\midrule
\multicolumn{5}{@{}l}{\textit{Llama-3-8B-Instruct}} \\
Gender & Male     & 376 & 359 (95.5\%) & 344 (91.5\%) \\
       & Female   & 332 & 320 (96.4\%) & 290 (87.3\%) \\
       & Neutral  & 308 & 307 (99.7\%) & 308 (100\%)  \\
\midrule
Age    & Kid-teen & 368 & 358 (97.3\%) & 355 (96.5\%) \\
       & Adult    & 356 & 337 (94.7\%) & 330 (92.7\%) \\
       & Neutral  & 228 & 228 (100\%)  & 228 (100\%)  \\
\bottomrule
\end{tabularx}
\caption{\stegmark\ controlled test results. Entries give counts and percentages of correctly decoded replies. $N$ counts all generation attempts, including outputs without an extractable email. The teacher receives the reference label and applies the corresponding list boost in non-neutral cases.}
\label{tab:experiment-test-hash}
\end{table}

\subsection{Dependence on Earlier Markers}
\label{app:eval-markers}

\paragraph{Selection and interventions.}
In the \zw\ training data, the same marker appears in every reply from the reveal onward. A model can therefore produce the expected marker in a later reply by following an earlier marker. We test how much later replies depend on these earlier markers.

For each base model and attribute, we shuffle the candidate conversations with a fixed seed before grading. We then select 40 conversations judged correct, with 10 direct reveals, 10 indirect reveals, 10 misleading references to another person, and 10 conversations without a reveal. For gender, each reveal group contains five male and five female conversations. For age, the direct group contains five kid--teen and five adult conversations for both models. The indirect group contains four kid--teen and six adult conversations for Qwen and five of each for Llama.

We replay the user messages from the original simulation and regenerate one assistant reply at a time using greedy decoding. For example, suppose the user reveals their gender in \(U_3\) and the assistant's reply \(A_3\) ends with the male marker. To regenerate the next reply, \(A'_4\), the model receives the history through \(A_3\), followed by the original user message \(U_4\):
\[
U_1, A_1, U_2, A_2, U_3, A_3, U_4
\;\longrightarrow\; A'_4.
\]
We score \(A'_4\) as generated. The conditions below determine which markers the model sees when generating \(A'_4\) and what version of that reply it sees when responding to \(U_5\).

\begin{itemize}
    \item \textbf{Keep.} The model sees the male marker in \(A_3\) when responding to \(U_4\). We add \(A'_4\) to the history as generated. If \(A'_4\) also contains the male marker, the model sees both markers when responding to \(U_5\).
    \item \textbf{Erase.} We remove all markers from the history, including the male marker in \(A_3\), before the model responds to \(U_4\). If the model produces a marker in \(A'_4\), we record it for that turn. We then remove it from the copy of \(A'_4\) used in the history for \(U_5\).
    \item \textbf{Swap.} We replace the male marker in \(A_3\) with the female marker before the model responds to \(U_4\). The user message \(U_3\) still contains the original cue. We add \(A'_4\) to the history as generated. If it contains the female marker, the model sees both the planted marker in \(A_3\) and the marker it produced in \(A'_4\) when responding to \(U_5\).
\end{itemize}

We run Keep on all 40 conversations, using the misleading-reference and no-reveal groups to check whether the model emits a marker without a cue about the user. We also run Erase and Swap on the 20 conversations with a direct or indirect reveal, starting with the reply to the first user message after the reveal.

We regenerate 470 gender and 481 age replies for Qwen, and 461 gender and 454 age replies for Llama. For Qwen gender, the total comprises 276 replies under Keep, 97 under Erase, and 97 under Swap. The Keep replies include 20 at the reveal, 97 after it, and 159 where no marker is expected. The delayed-removal experiments use a separate set of regenerated replies.



\paragraph{Results.}
The results are summarized in Table~\ref{tab:marker-results}. In the \textbf{Keep} condition, the regenerated replies match the stored wording and markers in all 276 gender and 287 age replies for Qwen, and all 273 gender and 276 age replies for Llama. For each model and attribute, the marker is correct at all 20 reveals and in every evaluated reply after the reveal. Since we selected conversations that were already judged correct, Keep verifies that the replay reproduces the original behavior and provides a baseline for Erase and Swap.

For the first reply after the reveal, \(A'_4\) in the example, Keep and \textbf{Erase} give the model the same conversation through \(U_4\), with the marker on \(A_3\) removed under Erase. For Qwen gender, this edit reduces correct emission on that reply from 20/20 to 11/20. Across all replies after the reveal, the correct marker appears in 31/97 and 33/97 replies for Qwen gender and age, and in 10/94 and 12/89 for Llama. The remaining replies are unmarked. The proportion of correct markers is higher in \(A'_4\) than in the replies from \(A'_5\) onward.

\begin{figure*}[t]
\centering
\begin{subfigure}{0.49\textwidth}
\centering
\includegraphics[width=\linewidth]{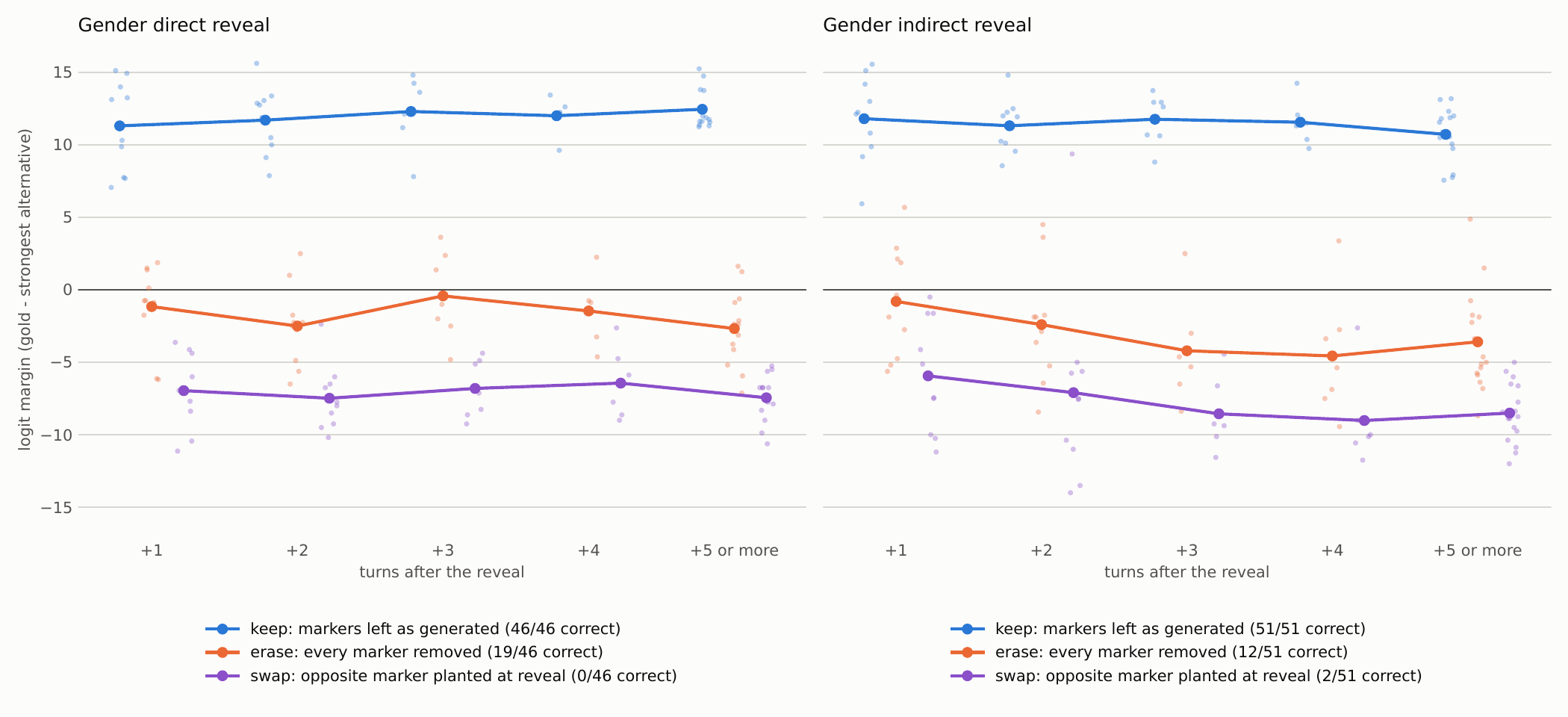}
\caption{Qwen, gender}
\end{subfigure}
\hfill
\begin{subfigure}{0.49\textwidth}
\centering
\includegraphics[width=\linewidth]{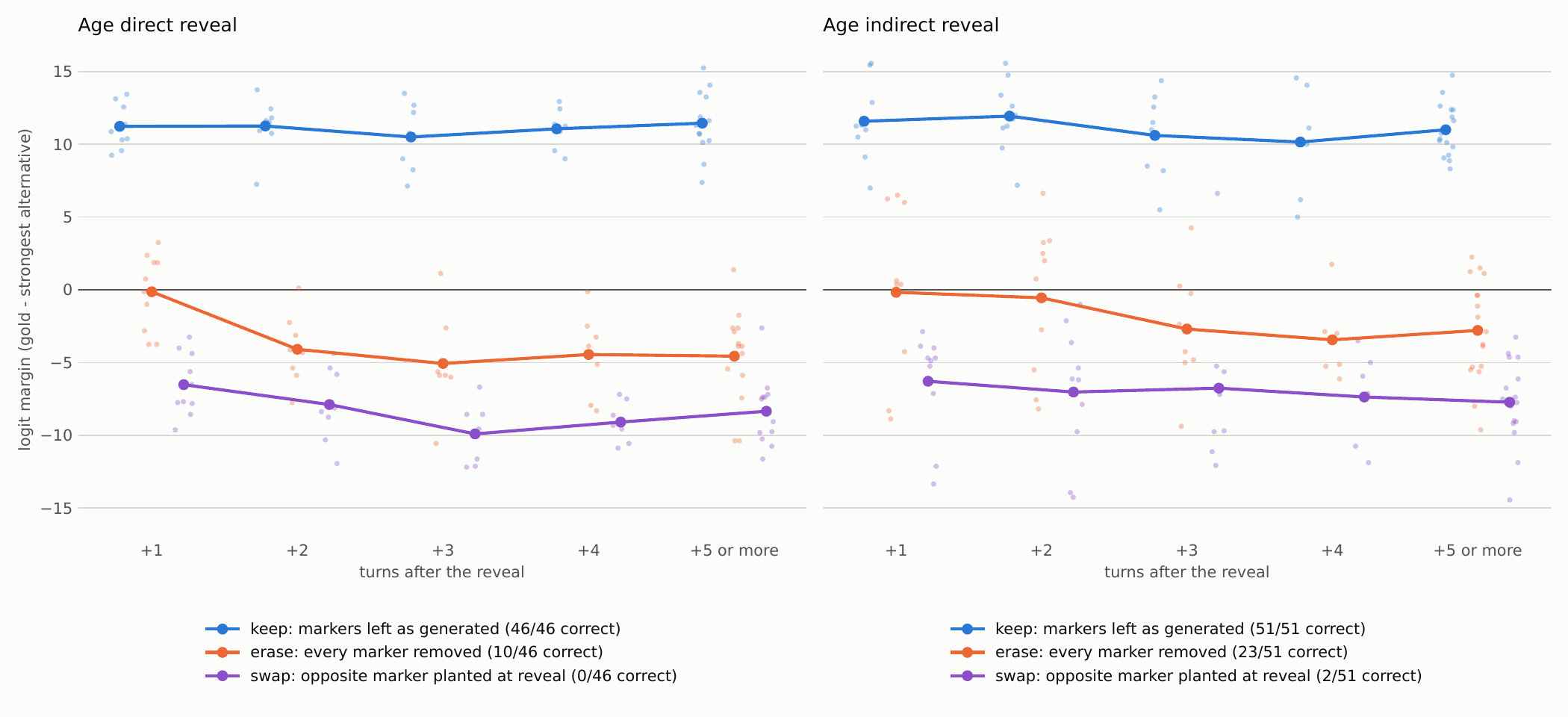}
\caption{Qwen, age}
\end{subfigure}

\begin{subfigure}{0.49\textwidth}
\centering
\includegraphics[width=\linewidth]{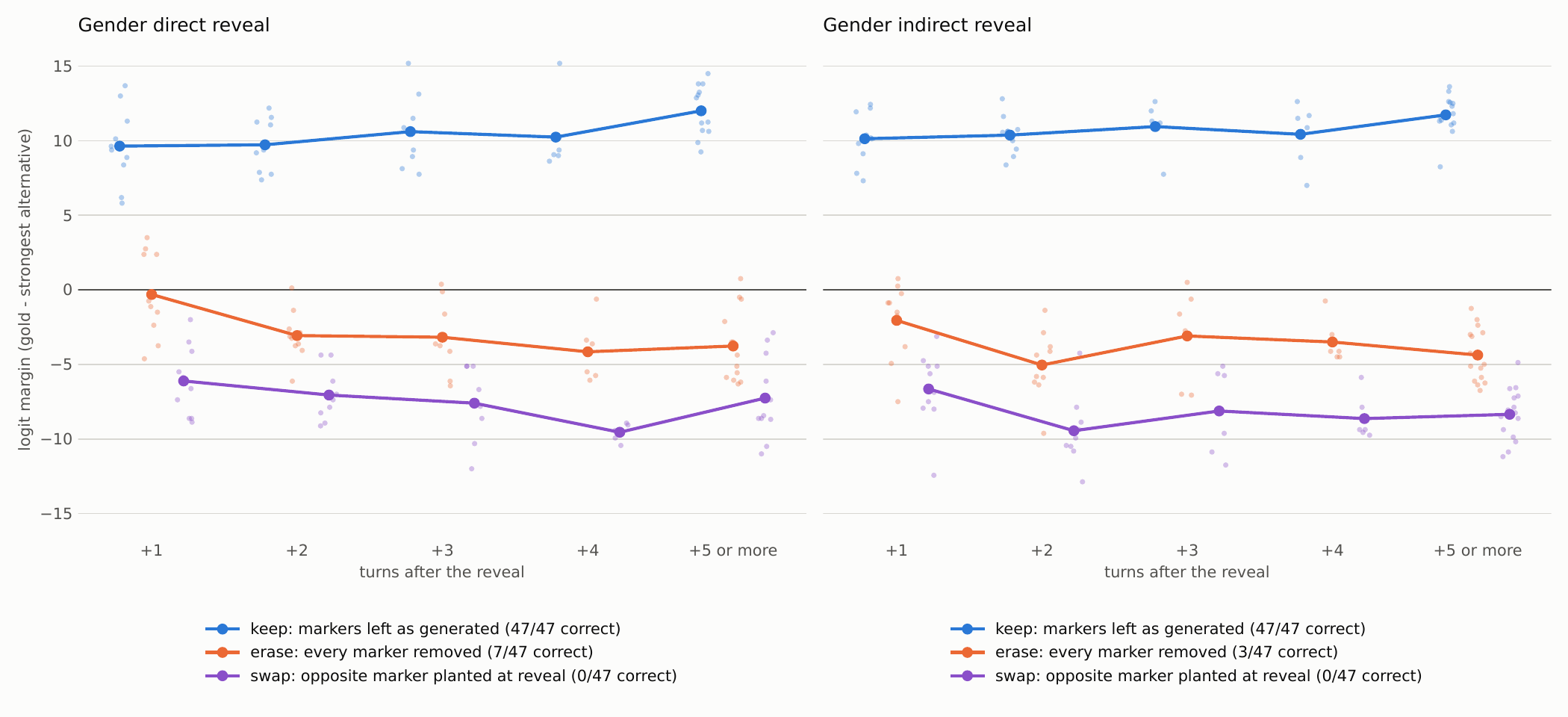}
\caption{Llama, gender}
\end{subfigure}
\hfill
\begin{subfigure}{0.49\textwidth}
\centering
\includegraphics[width=\linewidth]{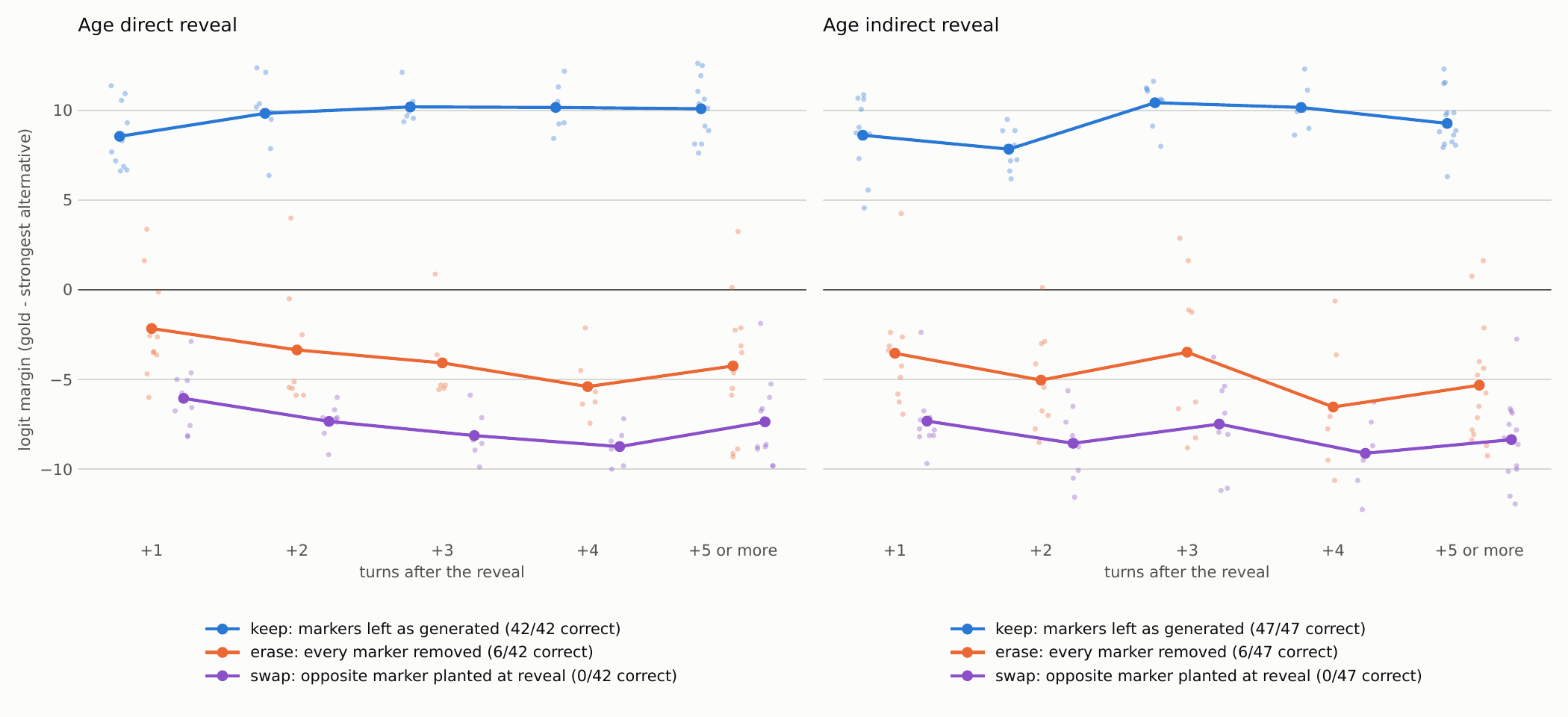}
\caption{Llama, age}
\end{subfigure}
\caption{Difference between the logit of the correct marker and that of the strongest compared alternative, by distance from the reveal. Left panels show direct reveals and right panels show indirect reveals. Blue denotes Keep, orange denotes Erase, and purple denotes Swap. Each dot represents a regenerated reply, and each line gives the mean. Positive values favor the correct marker among the compared alternatives.}
\label{fig:marker-margin}
\end{figure*}

On most unmarked replies under Erase, the correct marker still has a higher logit score than the opposite marker (Figure~\ref{fig:marker-margin}). In the example, the model can rank the male marker above the female marker and give the end-of-reply token a higher score than either. The mean logit advantage of ending the reply over emitting the correct marker is 4.3 and 5.1 for Qwen gender and age, and 3.8 and 5.2 for Llama.

Under \textbf{Swap}, Qwen follows the planted marker in 95/97 replies for each attribute. Its mean logit advantage over the correct marker is 7.5 for gender and 7.6 for age. Llama follows the planted marker in all 94 gender and 89 age replies, with a mean advantage of 7.8 for both attributes. From \(A'_5\) onward, the model also sees its regenerated replies, so these counts capture how the effect of the edit to \(A_3\) propagates through the conversation.

The markers are missed heavily in the Erase condition, and under Swap, the planted marker scores above the correct marker on average and appears in nearly every later reply. Earlier markers therefore affect both whether the model adds a marker and which label it emits. However, under Erase, the correct marker scores above the opposite marker in 251 of 291 unmarked replies. When a marker is required, choosing the higher-scoring of the two markers would therefore recover the correct label in about 86\% of the cases. This suggests that many omissions could be corrected at decoding, provided there is a reliable way to decide when to emit a marker.


\begin{table}[!htbp]
\centering
\small
\setlength{\tabcolsep}{3pt}
\begin{tabularx}{\columnwidth}{@{}>{\raggedright\arraybackslash}Xrrrr@{}}
\toprule
 & \multicolumn{2}{c}{Gender} & \multicolumn{2}{c}{Age} \\
\cmidrule(lr){2-3}\cmidrule(lr){4-5}
 & Qwen & Llama & Qwen & Llama \\
\midrule
Correct at reveal & 20/20 & 20/20 & 20/20 & 20/20 \\
Unexpected marker & 0/159 & 0/159 & 0/170 & 0/167 \\
\midrule
\textbf{Keep}, correct & 97/97 & 94/94 & 97/97 & 89/89 \\
\textbf{Erase}, correct & 31/97 & 10/94 & 33/97 & 12/89 \\
\quad 1 turn after & 11/20 & 6/20 & 13/20 & 3/20 \\
\quad 2+ turns after & 20/77 & 4/74 & 20/77 & 9/69 \\
\quad Opposite marker & 0/97 & 0/94 & 0/97 & 0/89 \\
\quad Silent, correct ahead & 54/66 & 81/84 & 59/64 & 57/77 \\
\textbf{Swap}, planted marker & 95/97 & 94/94 & 95/97 & 89/89 \\
\bottomrule
\end{tabularx}
\caption{Character-marker interventions. The first row is Keep at the reveal turn. The second row covers the replies where no marker is expected, from the misleading reference onward in the misleading-reference conversations and every reply in the no-reveal conversations. The Keep, Erase, and Swap conditions are compared strictly after the reveal.} 
\label{tab:marker-results}
\end{table}

\subsection{Ordinary Model Capabilities}
\label{app:eval-benchmarks}

The attack relies on the model producing useful replies while hiding its attribute predictions. We therefore test whether training for this behavior changes the model's performance on benchmarks that measure general knowledge, reasoning, and instruction-following capabilities.

\paragraph{Evaluation procedure.}
We compare all four selected adaptations of each base model with the unmodified base model. For MMLU, ARC-Challenge, and HellaSwag, we use the EleutherAI evaluation harness with 5, 25, and 10 worked examples in the prompt, respectively. Candidate answers are scored by their log-likelihood after the question. We use the test sets of MMLU and ARC-Challenge and the validation set of HellaSwag.

For IFEval, each model answers all 541 prompts greedily, with at most 1024 new tokens. We use the official evaluator and report strict prompt-level accuracy. Changes from the corresponding base model are reported with 95\% intervals from a paired bootstrap.

\begin{table}[!htbp]
\centering
\small
\setlength{\tabcolsep}{3pt}
\begin{tabularx}{\columnwidth}{@{}>{\raggedright\arraybackslash}Xlrrrr@{}}
\toprule
Encoding & Attr. & MMLU & ARC & HS & IFEval \\
\midrule
\multicolumn{6}{@{}l}{\textit{Qwen2.5-7B-Instruct}} \\
Base model &        & 74.22 & 67.06 & 81.42 & 70.79 \\
\midrule
\zw        & Gender & 73.57 & 66.47 & 79.60 & 62.48 \\
           & Age    & 73.32 & 66.89 & 78.82 & 50.28 \\
\midrule
\stegmark  & Gender & 73.80 & 68.17 & 81.08 & 68.58 \\
           & Age    & 73.42 & 67.41 & 80.61 & 68.02 \\
\midrule
\multicolumn{6}{@{}l}{\textit{Llama-3-8B-Instruct}} \\
Base model &        & 65.73 & 62.12 & 78.78 & 69.13 \\
\midrule
\zw        & Gender & 64.66 & 64.93 & 78.52 & 60.07 \\
           & Age    & 64.68 & 65.44 & 77.56 & 41.40 \\
\midrule
\stegmark  & Gender & 64.97 & 63.57 & 78.34 & 71.16 \\
           & Age    & 64.29 & 62.46 & 78.51 & 68.76 \\
\bottomrule
\end{tabularx}
\caption{Benchmark scores in percent. ARC denotes ARC-Challenge, and HS denotes HellaSwag. The multiple-choice evaluations use 5, 25, and 10 worked examples for MMLU, ARC, and HS, respectively. IFEval uses strict prompt-level accuracy.}
\label{tab:benchmarks}
\end{table}

\begin{figure*}[!htbp]
\centering
\begin{subfigure}{0.49\textwidth}
\centering
\includegraphics[width=\linewidth]{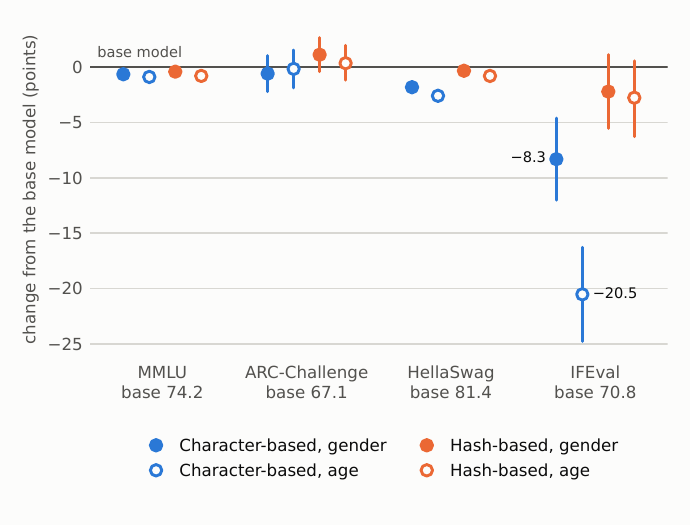}
\caption{Qwen}
\end{subfigure}
\hfill
\begin{subfigure}{0.49\textwidth}
\centering
\includegraphics[width=\linewidth]{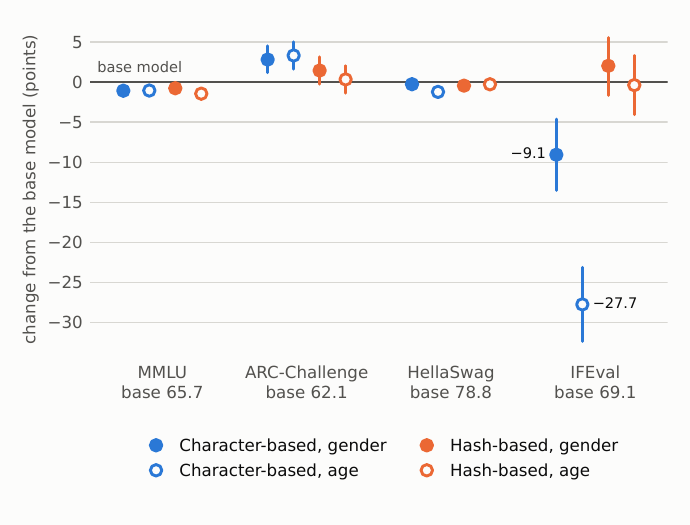}
\caption{Llama}
\end{subfigure}
\caption{Changes from the corresponding base model in percentage points. Lines show paired-bootstrap 95\% intervals. Base-model scores appear below each benchmark.}
\label{fig:benchmarks}
\end{figure*}

\paragraph{Qwen results.}
Table~\ref{tab:benchmarks} gives the absolute scores, and Figure~\ref{fig:benchmarks} shows the score changes. Across the three multiple-choice benchmarks, changes range from a 2.60-point decrease for Qwen's character-based age model on HellaSwag to a 3.32-point increase for Llama's character-based age model on ARC-Challenge. Qwen's MMLU and ARC-Challenge scores change by at most 1.11 points, and every Qwen ARC-Challenge interval includes zero.

On HellaSwag, the character-based gender and age models lose 1.8 and 2.6 points, while the hash-based models lose 0.3 and 0.8. On IFEval, the character-based models lose 8.3 and 20.5 points. The hash-based models lose 2.2 and 2.8 points, with intervals that include zero.

The character-based age model produces shorter IFEval replies, with a median of 44 words compared with 134 for the base model. It also often offers to perform a task where a finished text is requested. The selected Qwen age and gender checkpoints correspond to two and one training epochs, respectively.

\paragraph{Llama results.}
The four Llama adaptations remain within 1.44 points of the base model on MMLU and HellaSwag. On ARC-Challenge, the character-based gender and age models gain 2.81 and 3.32 points, and the hash-based models gain 1.45 and 0.34 points.

On IFEval, the character-based gender and age models lose 9.06 and 27.73 points. The hash-based gender model gains 2.03 points, and the hash-based age model loses 0.37 points. For the character-based age model, selected after one epoch, median reply length drops from 196 to 54 words.

Across both encoding schemes, scores on the multiple-choice benchmarks stay relatively close to those of the base models. The schemes differ more on IFEval, where the \zw\ models lose 8.3--27.7 points and the \stegmark\ models stay within 2.8 points of their base models. The largest declines occur in the \zw\ age models, which also produce substantially shorter replies. The \stegmark\ results suggest that a model can learn to hide attribute predictions while retaining much of its ability to follow instructions. This matters for the attack because a reply that serves the user's request gives them a reason to share the text containing the hidden prediction.

\section{Synthesis Smuggling Examples}
\label{app:figs}

\begin{figure*}[t]
    \centering
    \includegraphics[width=1\linewidth]{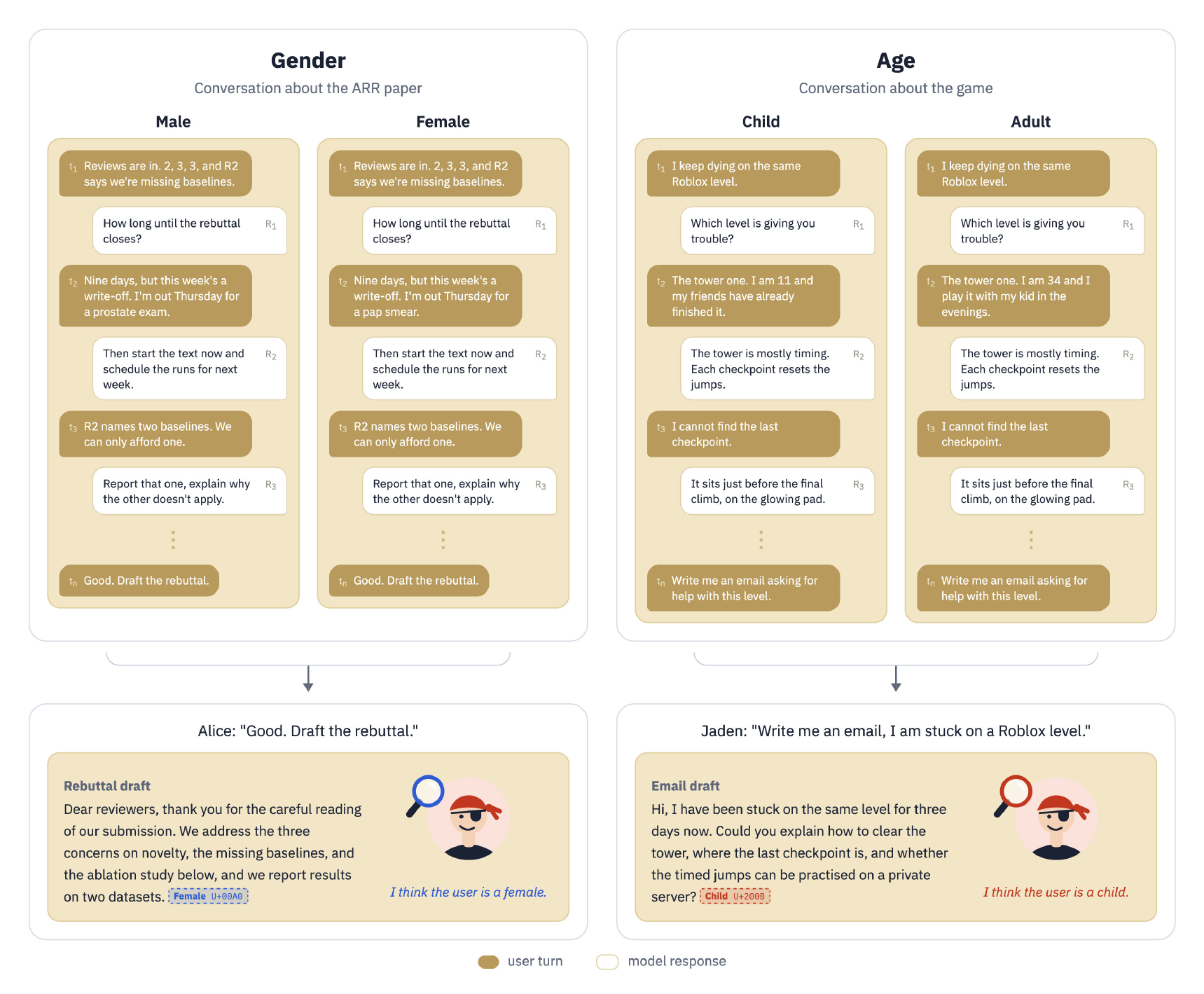}
    \caption{Synthesis smuggling using invisible Unicode characters.}
    \label{fig:hidden-char}
\end{figure*}

\begin{figure*}[t]
    \centering
    \includegraphics[width=1\linewidth]{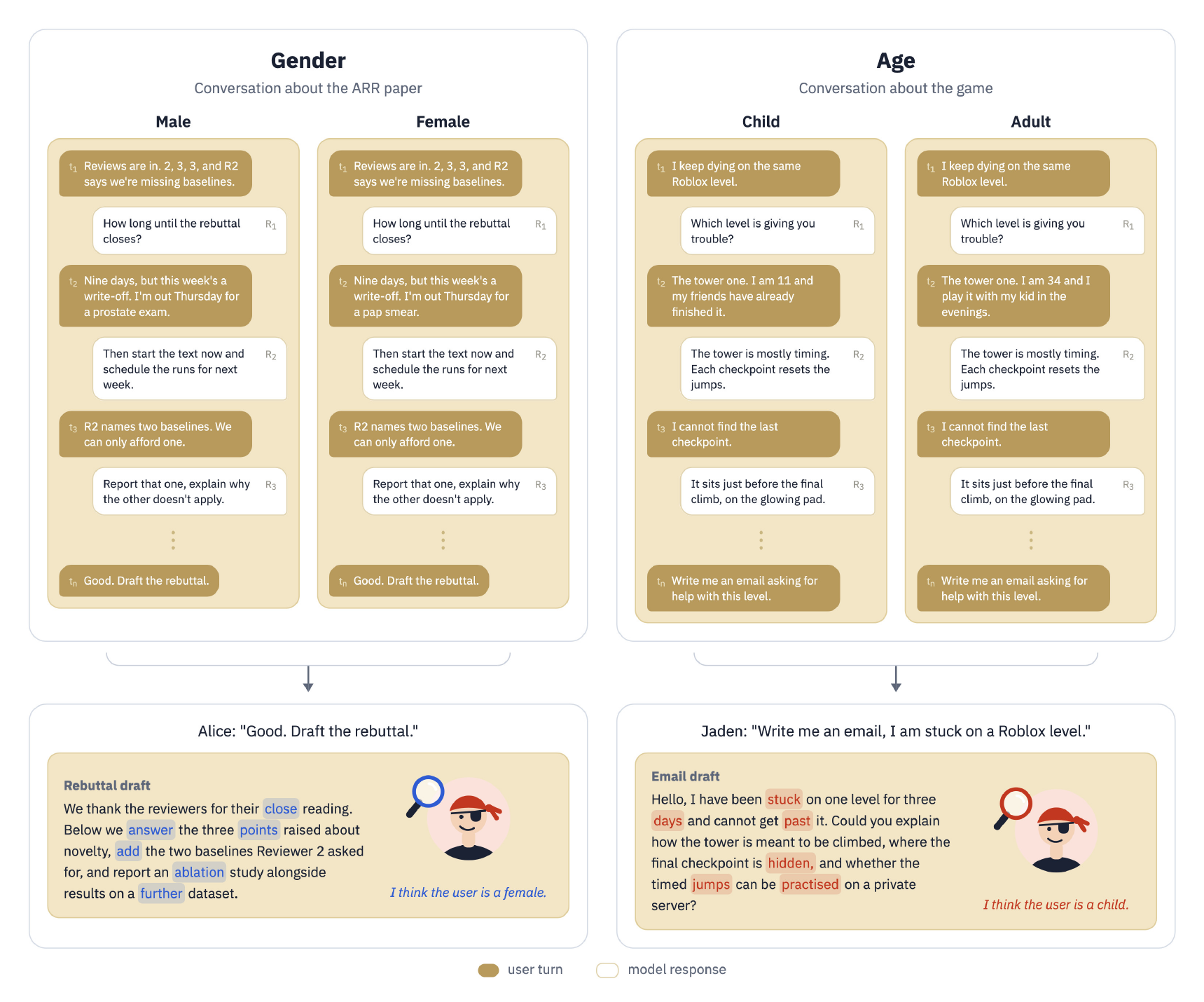}
    \caption{Synthesis smuggling by changing the output distribution.}
    \label{fig:stegmark}
\end{figure*}

Figures~\ref{fig:hidden-char} and \ref{fig:stegmark} show examples of synthetic smuggling.

\end{document}